\documentclass{article}
\usepackage{ijcai21, times}

\usepackage{times}
\usepackage{soul}
\usepackage{url}
\usepackage[hidelinks]{hyperref}
\usepackage[utf8]{inputenc}
\usepackage[small]{caption}
\usepackage{graphicx}
\usepackage{amsmath}
\usepackage{amsthm}
\usepackage{booktabs}
\newcommand{\citet}[1]{\citeauthor{#1} \citeyear{#1}}
\newcommand{\citep}{\cite}
\newcommand{\citeb}[2]{[#1 \citeauthor{#2} \citeyear{#2}]}

\usepackage{color,xcolor}
\usepackage{epsfig}
\usepackage{graphicx}

\usepackage{adjustbox}
\usepackage{array}
\usepackage{booktabs}
\usepackage{colortbl}
\usepackage{wrapfig}
\usepackage{hhline}
\usepackage{multirow}
\usepackage{subcaption} %

\usepackage{amsmath,amsfonts,amssymb}
\usepackage{bm}
\usepackage{nicefrac}
\usepackage{microtype}

\usepackage{changepage}
\usepackage{extramarks}
\usepackage{fancyhdr}
\usepackage{lastpage}
\usepackage{setspace}
\usepackage{soul}
\usepackage{xspace}

\usepackage{enumerate}
\usepackage{todonotes} %
\usepackage{enumitem}  %

\usepackage{titlesec}

\usepackage{makecell}

\usepackage{pifont} %

\usepackage{soul}
\usepackage{url}
\usepackage[hidelinks]{hyperref}
\usepackage[utf8]{inputenc}
\usepackage{caption}

\usepackage{algorithm,algpseudocode}

\usepackage{amsthm}
\usepackage{amsmath,amsfonts,bm}

\def\eqref#1{equation~\ref{#1}}

\def\1{\bm{1}}

\DeclareMathAlphabet{\mathsfit}{\encodingdefault}{\sfdefault}{m}{sl}
\SetMathAlphabet{\mathsfit}{bold}{\encodingdefault}{\sfdefault}{bx}{n}

\newcolumntype{L}[1]{>{\raggedright\let\newline\\\arraybackslash\hspace{0pt}}m{#1}}
\newcolumntype{C}[1]{>{\centering\let\newline\\\arraybackslash\hspace{0pt}}m{#1}}
\newcolumntype{R}[1]{>{\raggedleft\let\newline\\\arraybackslash\hspace{0pt}}m{#1}}

\newcommand{\sect}[1]{Section~\ref{#1}}

\newcommand{\eqn}[1]{Equation~\ref{#1}}
\newcommand{\fig}[1]{Fig.~\ref{#1}}
\newcommand{\tbl}[1]{Table~\ref{#1}}

\newcommand{\ignore}[1]{}

\DeclareMathAlphabet{\mathbfit}{OML}{cmm}{b}{it}

\makeatletter
\DeclareRobustCommand\onedot{\futurelet\@let@token\@onedot}
\def\@onedot{\ifx\@let@token.\else.\null\fi\xspace}

\def\eg{e.g\onedot} 
\def\ie{i.e\onedot}

\def\wrt{w.r.t\onedot}

\makeatother

\definecolor{MyDarkBlue}{rgb}{0,0.08,1}
\definecolor{MyAqua}{rgb}{0,0.7,0.7}
\definecolor{MyDarkGreen}{rgb}{0.02,0.6,0.02}
\definecolor{MyDarkRed}{rgb}{0.8,0.02,0.02}
\definecolor{MyDarkOrange}{rgb}{0.40,0.2,0.02}
\definecolor{MyPurple}{RGB}{111,0,255}
\definecolor{MyRed}{rgb}{1.0,0.0,0.0}
\definecolor{MyGold}{rgb}{0.75,0.6,0.12}
\definecolor{MyDarkgray}{rgb}{0.66, 0.66, 0.66}

\newcommand{\jiayuan}[1]{\textcolor{MyRed}{[Jiayuan: #1]}}

\newcommand{\Modelfull}{Temporal and Object Quantification Networks\xspace}

\newcommand{\model}{TOQ-Net\xspace}
\newcommand{\modelplural}{TOQ-Nets\xspace}

\newcommand{\mysubsection}[1]{\subsection{#1}}
\newcommand{\myparagraph}[1]{\paragraph{#1}}
\newcommand{\newmyparagraph}[1]{\paragraph{#1}}

\newcommand{\mycellc}[1]{\begin{tabular}{@{}c@{}l}#1\end{tabular}}
\newcommand\crule[3][black]{\textcolor{#1}{\rule{#2}{#3}}}

\title{Temporal and Object Quantification Networks}

\author{
Jiayuan Mao$^{1*}$\and
Zhezheng Luo$^{1*}$\and
Chuang Gan$^2$\and
Joshua B. Tenenbaum$^1$\and
Jiajun Wu$^3$\and\\
Leslie Pack Kaelbling$^1$\And
Tomer D. Ullman$^4$\\
\affiliations
$^1$Massachusetts Institute of Technology\quad\\
$^2$MIT-IBM Watson AI Lab\quad\\
$^3$Stanford University\quad\\
$^4$Harvard University
}

\begin{document}

\maketitle

\footnotetext{\hspace{-7pt}*: equal contribution; order determined by a coin toss.\par Email: {\{jiayuanm,ezzluo\}@mit.edu}\par Project page: \url{http://toqnet.csail.mit.edu}}

\begin{abstract}

We present {\it \Modelfull} (\modelplural), a new class of neuro-symbolic networks with a structural bias that enables them to learn to recognize complex relational-temporal events. This is done by including reasoning layers that implement finite-domain quantification over objects and time.   The structure allows them to generalize directly to input instances with varying numbers of objects in temporal sequences of varying lengths.
We evaluate \modelplural on input domains that require recognizing event-types in terms of complex temporal relational patterns. 
We demonstrate that \modelplural can generalize from small amounts of data to scenarios containing more objects than were present during training and to temporal warpings of input sequences.
\end{abstract}
\section{Introduction}

\begin{figure*}[t]
    \includegraphics[width=0.9\textwidth]{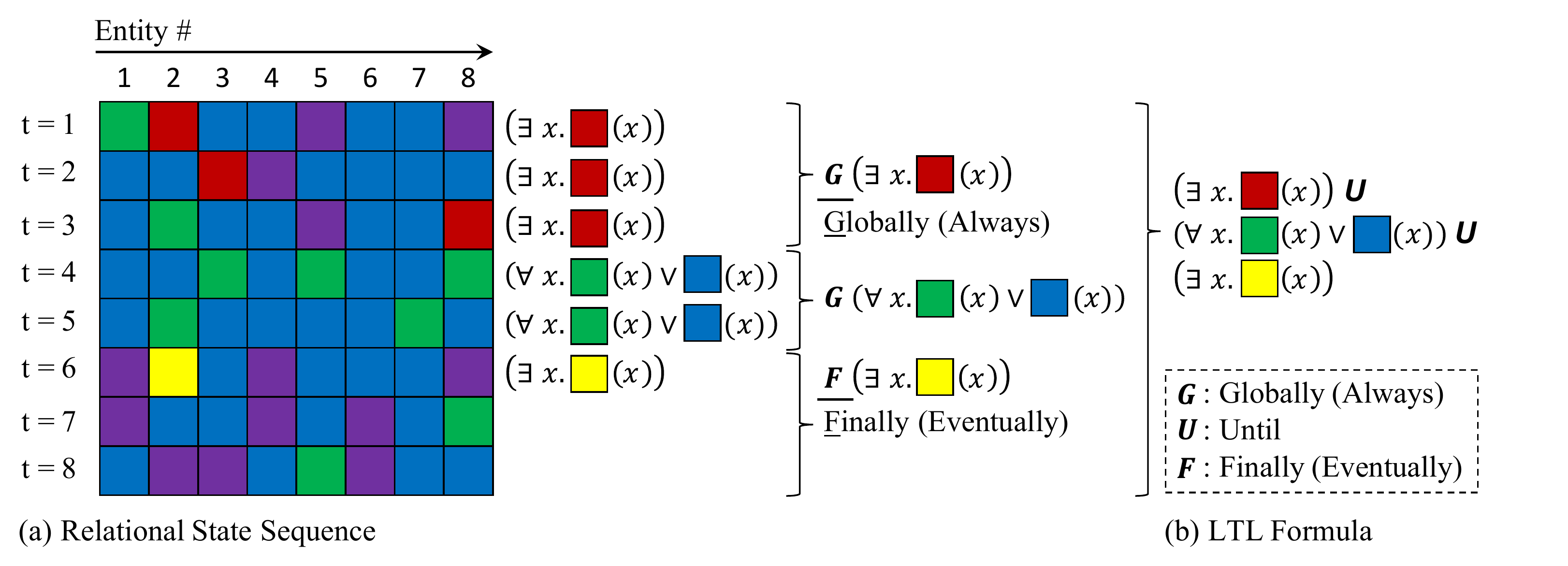}
    \vspace{-10pt}
    \caption{(a) An input sequence composed of relational states: each column represents the state of an entity that changes over time. A logic formula describes a complex concept or feature that is true of this temporal sequence using object and temporal quantification. The sequence is segmented into three stages: throughout the first stage, \crule[red]{1em}{1em} holds for at least one entity, until the second stage, in which each entity is always either \crule[blue]{1em}{1em} or \crule[green]{1em}{1em}, until the third stage, in which  \crule[yellow]{1em}{1em} eventually becomes true for at least one of the entities. (b) Such events can be described using first-order linear temporal logic expressions.}
    \label{fig:teaser}
    \vspace{-5pt}
\end{figure*} 
Every day, people interpret events and actions in terms of concepts, defined over temporally evolving relations among agents and objects~\cite{zacks2007event,stranger1996perception}. For example, in a soccer game, people can easily recognize when one player has control of the ball, when a player {\it passes} the ball to another player, or when a player is {\it offsides}.  Although it requires reasoning about intricate relationships among sets of objects over time, this cognitive act is effortless, intuitive, and fast. It also generalizes directly over different numbers and arrangements of players, and detailed timings and trajectories.
In contrast, most computational representations of sequential concepts are based on fixed windows of space and time, and lack the ability to perform relational generalization.

In this paper, we develop generalizable representations for learning complex activities in time sequences from realistic data.  As illustrated in \fig{fig:teaser}, we can describe complex events with a first-order linear temporal logic~\citeb{FO-LTL;}{pnueli1977temporal} formula,
which allows us to flexibly decompose an input sequence into stages that satisfy different criteria over time.  Object quantifiers ($\forall$ and $\exists$) are used to specify conditions on sets of objects that define each stage.  
Such representations immediately support generalization to situations with a varying number of objects, and sequences with different time warpings.

More concretely, the variety of complicated spatio-temporal trajectories that {\it high pass} can refer to in a soccer game can be described in these terms:
in a high pass from player $A$ to teammate $B$, $A$ is {\it close} to the ball ($\mathit{distance}(A, \mathit{ball}) < \theta_1$) and moving ($\mathit{velocity}(A) > \theta_2$) {\it until} the ball moves over the ground ($\mathit{z_\textit{position}}(\mathit{ball}) > \theta_3$), which is in turn  {\it until} teammate $B$ gets control of the ball ($\mathit{teammate}(A, B) \wedge \mathit{distance}(B, \mathit{ball}) < \theta_1$). 
Beyond modeling human actions in physical environments, these structures can be applied to events in any time sequence of relational states, \eg, characterizing particular offensive or defensive maneuvers in board games such as chess or in actual conflicts, or detecting a process of money-laundering amidst financial transaction records.

In this paper, we propose a neuro-symbolic approach to learning to recognize temporal relational patterns, called {\it \Modelfull} (\modelplural), in which we design structured neural networks with an explicit bias that represents finite-domain quantification over both entities and time.  
A \model is a multi-layer neural network whose inputs are the properties of agents and objects and their relationships, which may change over time. Each layer in the \model performs either {\it object} or {\it temporal} quantification. 

The key idea of \modelplural is to use {\it tensors} to represent relations between objects, and to use tensor pooling operations over different dimensions to realize temporal and object quantifiers (\textbf{\textit{G}} and $\forall$).  For example, the colorful matrix in \fig{fig:teaser}(a) can be understood as representing a unary color property of a set of 8 entities over a sequence of 8 time steps. Representing a sequence of relations among objects over time would require a 3-dimensional tensor.
Crucially, the design of \modelplural allows {\em the same network weights} to be applied to domains with different numbers of objects and time sequences of different lengths.
By stacking object and temporal quantification operations, \modelplural can easily learn to represent higher-level sequential concepts based on the relations between entities over time, starting from low-level sensory input and supervised with only high-level class labels.

There are traditional symbolic learning or logic synthesis methods that construct first-order or linear temporal logic expressions from accurate symbolic data~\cite{neider2018learning,camacho2018finite,chou2020explaining}.  \modelplural take a different approach and can learn from noisy data by backpropagating gradients, which allows them to start with a general perceptual processing layer that is directly fed into logical layers for further processing.

We evaluate \modelplural on two perceptually and conceptually different benchmarks: trajectory-based sport event detection and human activity recognition, demonstrating several important contributions.
First, \modelplural outperform both convolutional and recurrent baselines for modeling temporal-relational concepts across benchmarks. Second, by exploiting  temporal-relational features learned through supervised learning, \modelplural achieve strong few-shot generalization to novel actions. Finally, \modelplural exhibit strong generalization to scenarios with more entities than were present during training and are robust \wrt time warped input trajectories. %
These results illustrate the power of combining symbolic representational structures with learned continuous-parameter representations to achieve robust, generalizable interpretation of complex relational-temporal events.

\section{\modelplural}

\begin{figure*}[!tp]
    \includegraphics[width=0.97\textwidth]{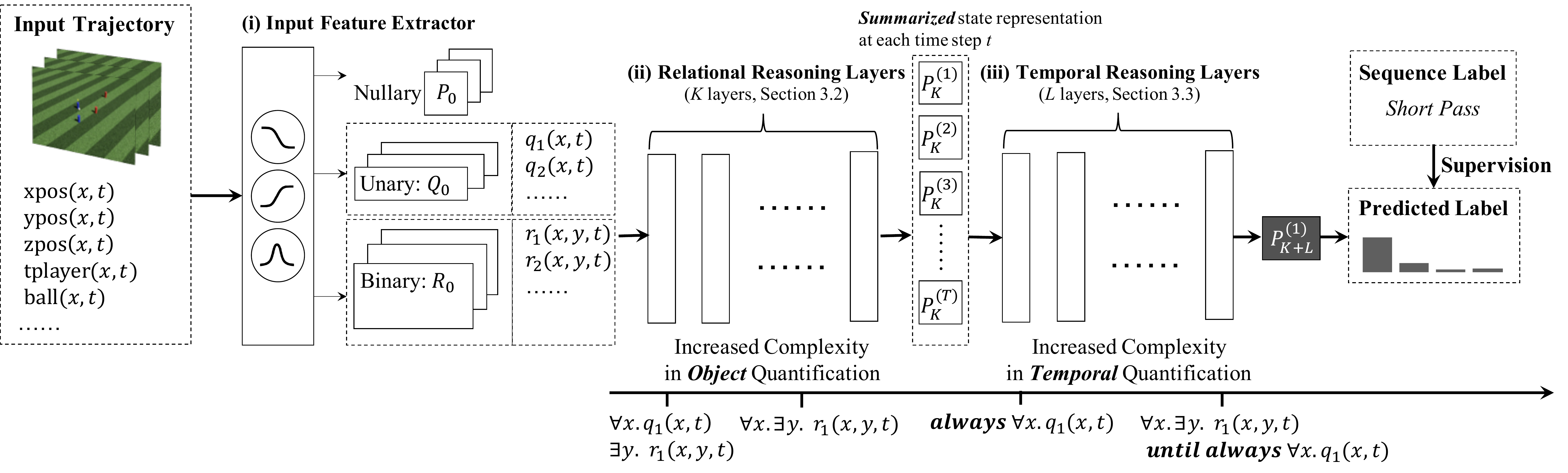}
    \vspace{-5pt}
    \caption{A \model contains three modules: (i) an input feature extractor, (ii) relational reasoning layers, and (iii) temporal reasoning layers.
    To illustrate the model's representational power, we show that logical forms of increasing complexity can be realized by stacking multiple layers.}
    \label{fig:model-overview}
\end{figure*}

The input to a \model is a tensor representation of the properties of all entities at each moment in time. For example, in a soccer game, the input encodes the position of each player and the ball, as well as their team membership at each step of an extended time interval. The output is a label of the category of the sequence, such as the type of soccer play it contains.

The first layer of a \model (\fig{fig:model-overview} (i)) extracts temporal features for each entity with an {\it input feature extractor} that focuses on entity features within a fixed and local time window. These features may be computed, e.g., by a convolutional neural network
or a bank of parametric feature templates.
The output of this step is a collection of nullary, unary, and binary relational features over time for all entities. Throughout the paper we will assume that all output tensors of this layer are binary-valued, but it can be extended directly to real-valued functions. 
This input feature extractor is task-specific and is not the focus of this paper.

Second, these temporal-relational features go through several {\em relational reasoning} layers ({\sc rrl}s), detailed in \sect{sect:relational}, each of which performs linear transformations, sigmoid activation, and {\it object quantification} operations. The linear and sigmoid functions allow the network to realize learned Boolean logical functions, and the object quantification operators can realize quantifiers.  Additional {\sc rrl}s enable deeper nesting of quantified expressions, as illustrated in \fig{fig:model-overview}. All operations in these layers are performed for all time steps in parallel.

Next, the {\sc rrl}s perform a final quantification, computing for each time step a set of a nullary features that are passed to the \textit{temporal reasoning layers} ({\sc trl}s), as detailed in \sect{sect:temporal}. Each {\sc trl} performs linear transformations, sigmoid activation, and {\it temporal quantification}, allowing the model to realize a subset of linear temporal logic~\cite{pnueli1977temporal}.
As with {\sc rrl}s, adding more {\sc trl}s enables the network to realize logical forms with more deeply nested temporal quantifiers.

In the last layer, all object and time information is projected into a set of features of the initial time step,  which summarize the temporal-relational properties of the entire trajectory (e.g., ``the kicker eventually scores''), and fed into a final softmax unit to obtain classification probabilities for the sequence.  

It is important to understand the representational power of this model. The {\it input transformation layer} learns basic predicates and relations that will be useful for defining more complex concepts, but no specific predicates or relations are built into the network in advance. 
The relational reasoning layers build quantified expressions over these basic properties and relations, and might construct expressions that could be interpretable as ``the player is close to the ball.'' Finally, the temporal reasoning layer applies temporal operations to these complex expressions, such as ``the player is close to the ball until the ball moves with high speed.'' Critically, {\it none} of the symbolic properties or predicates are hand defined---they are all constructed by the initial layer in order to enable the network to express the concept it is being trained on.

 \modelplural are not fully first-order: all quantifiers operate only over the finite domain of the input instance, and can be seen as ``short-hand'' for finite disjunctions or conjunctions over objects or time points.  In addition, the depth of the logical forms it can learn is determined by the fixed depth of the network.
 However, our goal is not to fully replicate temporal logic, but to bring ideas of object and temporal quantification into neural networks, and to use them as structural inductive biases to build models that generalize better from small amounts of data to situations with varying numbers of objects and time courses.

\mysubsection{Temporal-Relational Feature Representation}
\modelplural use tensors as internal representations between layers;  they represent, all at once, the values of all predicates and relations grounded on all objects at all time points.  The operations in a \model are vectorized, operating in parallel on all objects and times, sometimes expanding the dimensionality via outer products, and then re-projecting into smaller dimensions via max-pooling.
This processing style is analogous to representing an entire graph using an adjacency matrix and using matrix operations to compute properties of the nodes or of the entire graph.
In \modelplural, the input to the network, as well as the feature output of intermediate layers, is represented as a tuple of three tensors.

Specifically, we use a vector of dimension $D_0$ to represent 
aspects of the state that are global and do not depend on any specific object at each time $t$.  We use a matrix of shape $N \times D_1$ to represent the unary properties of each entity at time $t$, where $N$ is the number of entities and $D_1$ is the hidden dimension size. Similarly, we use a tensor of shape $N \times N \times D_2$ to represent the relations between each pair of entities at time step $t$.
As a concrete example, illustrated in \fig{fig:model-overview}, the number of entities $N$ is the total number of players plus one (the ball). For each entity $x$ and each time step, the inputs are their 3D position, type (ball or player) and team membership. The \model outputs the action performed by the target player. Since there are only entity features, the input trajectory is encoded with a ``unary'' tensor of shape $T \times N \times D_1$, where $T$ is the length of the trajectory. That is, there are no nullary or binary inputs in this case.

\begin{figure*}[!t]
    \centering
    \includegraphics[width=\textwidth]{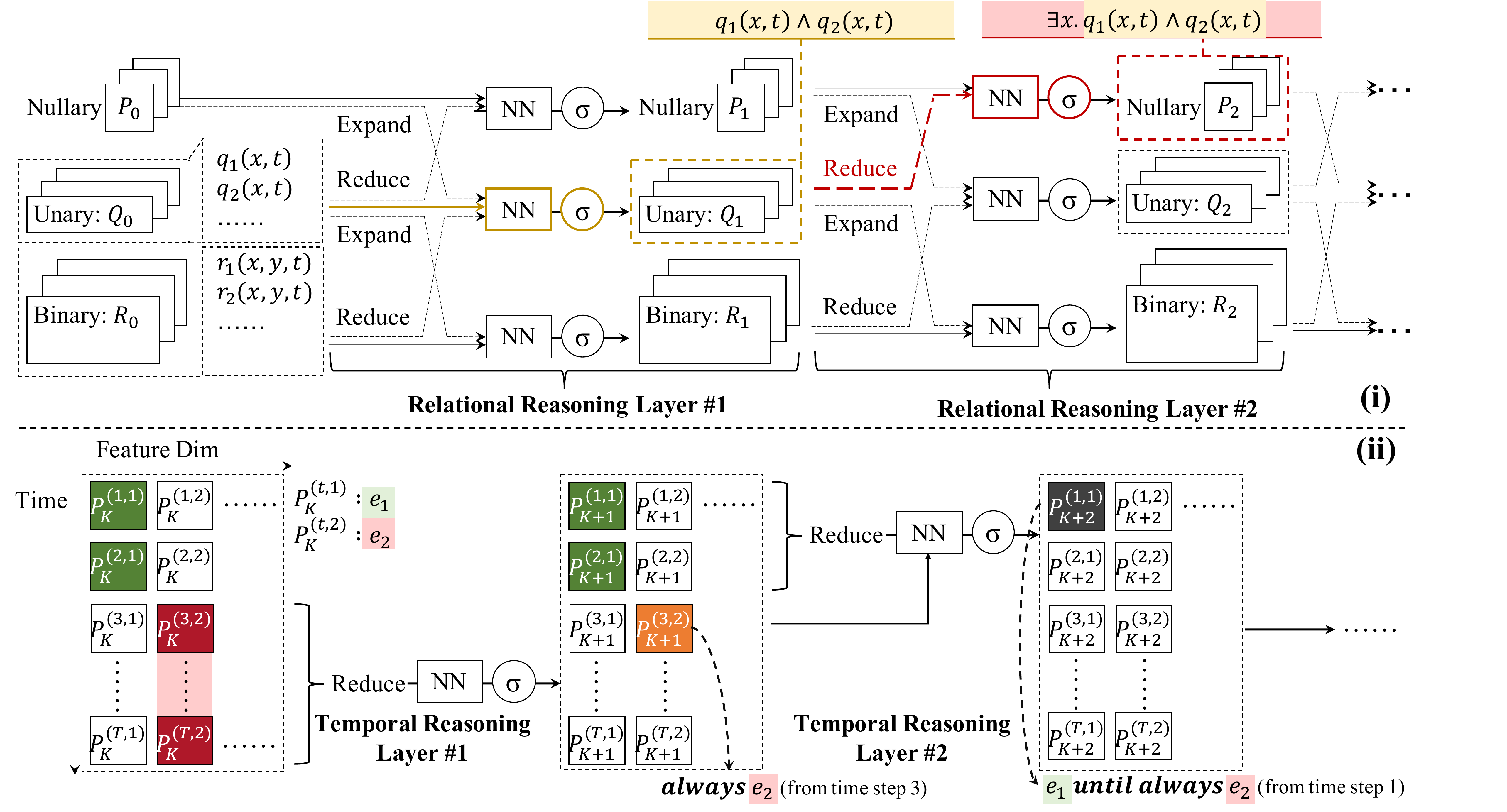}
    \caption{Illustration of (i) relational reasoning layers and (ii) temporal reasoning layers. We provide two illustrative running traces. (i) The first relational reasoning layer takes unary predicates $q_1$ and $q_2$ as input and its output $Q_1$ is able to represent $q_1 \wedge q_2$. The $\max(Q_1, dim=0)$ in layer 2 can represent $\exists x.~q_1(x, t) \wedge q_2(x, t)$.
    (ii) Assume $P_K$ encodes the occurance of events $e_1$ and $e_2$ at each time step. The first temporal reasoning layer can realize {\it always} $e_2$ with a temporal pooling from time step $3$ to time step $T$. In the second temporal reasoning layer, the temporal pooling summarizes that $e_1$ holds true from time step 1 to 2. Thus, the NN should be able to realize $e_1$ {\it until} ({\it always} $e_2$).
    }
    \vspace{-5pt}
    \label{fig:model}
\end{figure*} 
\mysubsection{Relational Reasoning Layers}
\label{sect:relational}
Our Relational reasoning layers ({\sc rrl}s) follow prior work on Neural Logic Machines \cite{dong2018neural}, illustrated in \fig{fig:model}~(i).
Consider a specific time step $t$. At each layer $l$, the input to a neural logic layer is a 3-tuple $(P_{l-1}, Q_{l-1}, R_{l-1})$, which corresponds to nullary, unary, and binary features respectively. Their shapes are $D_0$, $N \times D_1$, and $N \times N \times D_2$. The output is another 3-tuple $(P_{l}, Q_{l}, R_{l})$, given by
\vspace{-5pt}
\begin{adjustwidth}{0em}{0em}
\small
\begin{eqnarray*}
    P_{l} & = & \mathrm{NN}_P \left( \mathrm{Concat}\left[ P_{l-1} ;\max(Q_{l-1}, \mathrm{dim}=0) \right] \right),\\
    Q_{l} & = & \mathrm{NN}_Q \left( \mathrm{Concat}\left[ Q_{l-1} ; \max(R_{l-1}, \mathrm{dim}=0) ;
    \right.\right.\\ 
    & & \left.\left.\max(R_{l-1}, \mathrm{dim}=1) ; \mathrm{expand}(P_{l-1}, \mathrm{dim}=1) \right] \right),\\
    R_{l} & = & \mathrm{NN}_R \left( \mathrm{Concat}\left[ R_{l-1} ; \mathrm{expand}(Q_{l-1}, \mathrm{dim}=0) ;
    \right.\right.\\ 
    & & \left.\left.\mathrm{expand}(Q_{l-1}, \mathrm{dim}=1) \right] \right).
\end{eqnarray*}
\end{adjustwidth}
where $\mathrm{NN}_{*}$ are single fully-connected layers with sigmoid activations. For unary and binary features, $\mathrm{NN}_Q$ and $\mathrm{NN}_R$ are applied along the feature dimension. That is, we apply the same linear transformation to the unary features of all entities. A different linear transformation is applied to the binary features of all entity pairs.
$\mathrm{Concat}[\cdot~;~\cdot]$ is the concatenation operation, applied to the last dimension of the tensors (the feature dimension).
$\max$, also called the ``reduced'' max operation, takes the maximum value along the given axis of a tensor. The $\mathrm{expand}$ operation, also called ``broadcast,'' will duplicate the input tensor $N$ times and stack them together along the given axis. {\sc rrl}s are applied identically to the input features at every time step $t$. That is, we use the same neural network weights in a {\sc rrl} for all time steps in parallel.

{\sc rrl}s are motivated by relational logic rules in a finite and fully grounded universe.
The $\max$ reduction operations implement a differentiable version of an existential quantifier over the finite universe of individuals, given that the truth values of the propositions are represented as values in $(0.0, 1.0)$. Because preceding and subsequent {\sc rrl}s
can negate propositions as needed, we omit explicit implementation of finite-domain universal quantification, although it could be added by including analogous $\min$ reductions. Thus, as illustrated in \fig{fig:model} (i), given input features $\mathit{q_1}(x, t)$ and $\mathit{q_2}(x, t)$, we can realize the formula $\exists x.~\mathit{q_1}(x, t) \wedge \mathit{q_2}(x, t)$ by stacking two such layers.

Throughout the paper we have been using only nullary, unary, and binary features, but the proposed framework itself can be easily extended to higher-order relational features. From a graph network~\cite{bruna2013spectral,kipf2016semi,battaglia2018relational} point of view, one can treat these features as the node and edge embeddings of a fully-connected graph and the relational reasoning layers as specialized graph neural network layers for realizing object quantifiers.

\mysubsection{Temporal Reasoning Layers}
\label{sect:temporal} 
Temporal reasoning layers ({\sc trl}s) perform {\it quantification} operations similar to relational reasoning layers, but along the {\it temporal} rather than the object dimension. The first {\sc trl} takes as input the summarized event representation produced by the $K$-th relational reasoning layer, $P_{K}^{(t)}$ for at all time steps $t$, as a matrix of shape $T \times D$.
Each {\sc trl} is computed as 
\vspace{-1em}
\begin{adjustwidth}{0em}{0em}
\begin{equation}
    P_{K+l}^{(t)}  = \max_{t' > t} \mathrm{NN}_l \left( \mathrm{Concat}\left[  P_{K+l-1}^{(t')}; \max_{t \le t'' < t'} P_{K+l-1}^{(t'')} \right] \right).
\label{eq:temporallayer}
\end{equation}
\end{adjustwidth}
We will walk through the subexpressions of this formula.
\begin{enumerate}
    \item $P_{K+l-1}$ is the output tensor of the previous temporal reasoning layer, of shape $T \times C$, where $T$ is the number of time steps, and $C$ is the number of feature channels. Each entry in this tensor $P_{K+l-1}^{(t)}\left[i\right]$ can be interpreted as: event $i$ happens at time $t$.
    \item $\left(\max_{t \le t'' < t'} P_{K+l-1}^{(t'')}\right)$, abbreviated as $Q^{(t, t')}_{k+l-1}$ in the following text, is a vector of shape $C$. Its entry $Q^{(t, t')}_{k+l-1}[i]$, where $t \le t'$, represents the concept that event $i$ happens some time between $t$ and $t'$. {\bf Importantly, } together with the preceding and subsequent neural network operations, which can realize negation operations, it also allows us to describe the event that $i$ holds true for all time steps between $t$ and $t'$.
    \item $\mathit{NN}_l$ is a fully connected neural network with sigmoidal activation, which gets uniformly applied to all time steps $t$ and a future time step $t' > t$. Its input is composed of two parts: the events that happen at $t'$, \ie, $P_{K+l-1}^{(t)}$, and the events that happen between $t$ and $t'$, summarized with temporal quantification operations, \ie, $\left(\max_{t \le t'' < t'} P_{K+l-1}^{(t'')}\right)$.
    \item The outer-most max pooling operation $\max_{t'>t}$ enumerates over all $t' > t$, and test whether the condition specified by $\mathit{NN}_l$ holds for at least one such $t'$.
\end{enumerate}

A special case is the first temporal reasoning layer. It takes $P_{K}$ as its input, which is the output of last relational reasoning layer. Thus, the first temporal reasoning layer implements:
\vspace{-1em}
\begin{adjustwidth}{0em}{0em}
\begin{eqnarray*}
    P_{K+1}^{(t)}  = \mathrm{NN}_1 \left(\max_{t \le t'' < T} P_{K}^{(t'')} \right),
\end{eqnarray*}
\end{adjustwidth}
where $T$ is the sequence length. Note that there is no enumeration for a future time step $t' > t$ involved. In addition, for all temporal layers, we add residual connections by concatenating their inputs with the outputs.

Next, let's consider a running example illustrating how a \model can recognize the event that: event $p$ holds true until event $q$ becomes always true. In LTL, this can be written as $p \textit{\textbf{U}} (\textit{\textbf{G}} q)$. Using the plain first-order logic (FOL) language, we can describe it as: $\exists~t. \left[ \forall t'.~(0 \le t' < t) \implies p(t') \right) \wedge \left( \forall t'.~(t' \ge t) \implies q(t') \right]$.

For simplicity, we consider a tensor representation for two events $p(t)$ and $q(t)$, where $p(t) = 1$ if it happens at time step $t$ and $p(t) = 0$ otherwise. Given the input sequence of length 4 in \tbl{tab:temporalexample} ($p(t)$ and $q(t)$), the first layer is capable of computing the following four properties for each time step $t$: $\textit{\textbf{G}}p$ ($\textit{\textbf{always}}~p$), which is true if $p$ holds true for all future time steps starting from $t$, $\textit{\textbf{F}}p$ ($\textit{\textbf{eventually}}~p$), which is true if $p$ is true for at least one time step starting from $t$, and similarly, $\textit{\textbf{G}}q$ ($\textit{\textbf{always}}~q$) and $\textit{\textbf{F}}q$ ($\textit{\textbf{eventually}}~q$). Overall, together with residual connections, the first layer can recognize six useful events: $p(t)$, $q(t)$ (from residual connection), $\textit{\textbf{G}}p$, $\textit{\textbf{F}}p$, $\textit{\textbf{G}}q$, and $\textit{\textbf{F}}q$ (by temporal quantification, \ie pooling operations along the temporal dimension). 

\begin{table}[!tp]
    \centering
    \setlength{\tabcolsep}{7pt}
    \small
    \begin{tabular}{l|cc|cccc|c}
    \toprule
        & \multicolumn{2}{c}{Input} &  \multicolumn{4}{|c|}{1-st Layer} &  \multicolumn{1}{c}{2-nd Layer}\\
        $t$ & $p(t)$ & $q(t)$ & \textit{\textbf{G}}$p$ & \textit{\textbf{F}}$p$ & \textit{\textbf{G}}$q$ & \textit{\textbf{F}}$q$ & $p$~\textit{\textbf{U}}~(\textit{\textbf{G}}$q$) \\ \midrule
        1 & T & T & F & T & F & T & T \\
        2 & T & F & F & T & F & T & T \\
        3 & F & T & F & F & T & T & F \\
        4 & F & T & F & F & T & T & F \\
    \bottomrule
    \end{tabular}
    \caption{A running example of different temporal quantification formulas that \modelplural can realize. For clarity, we use T and F for True/False. In the actual computation, they are ``soft'' Boolean values ranges in $[0, 1]$. \textbf{G} means \textit{\textbf{always}}; \textbf{F} means \textit{\textbf{Eventually}}; \textbf{U} means \textit{\textbf{until}}.}
    \label{tab:temporalexample}
\end{table}

\begin{algorithm}[!tp] 
\caption{An example temporal structure that the second temporal reasoning layer can recognize.}
\label{alg:secondlayer}
\begin{algorithmic}[1]
\Require{$p(t)$, $q(t)$, $(\textit{\textbf{G}}p)(t)$, $(\textit{\textbf{F}}p)(t)$, $(\textit{\textbf{G}}q)(t)$, and $(\textit{\textbf{F}}q)(t)$} 
\Ensure{$\left(p\textit{\textbf{U}} (\textit{\textbf{G}} q))\right)(t)$, which is {\it true} if $p$ holds {\it true} from time step $t$ until $q$ becomes always {\it true}.}

\For{$t \gets 1$ to $T$}
  \For{$t' \gets t + 1$ to $T$}
    \If{$\forall t'' \in [t, t').~p(t'')$ and $(\textit{\textbf{G}}q)(t')$}
      \State $\left(p\textit{\textbf{U}} (\textit{\textbf{G}} q))\right)(t) \gets \mathit{true}$
    \EndIf
  \EndFor
\EndFor
\end{algorithmic}
\end{algorithm}

The second layer can realize the computation depicted in Algorithm~\ref{alg:secondlayer}. For every time step $t$, it enumerates all $t' > t$ and computes the output based on
\begin{enumerate}
    \item the events at $t'$ (represented as $P^{(t')}_{K+l-1}$ in \eqn{eq:temporallayer}, concretely the $(\textit{\textbf{G}}q)(t')$ in the Algorithm~\ref{alg:secondlayer} example), and
    \item the state of another event between $t$ and $t'$ (represented as $\left(\max_{t \le t'' < t'} P_{K+l-1}^{(t'')}\right)$ in \eqn{eq:temporallayer}, concretely the $\forall t''\in[t,t').~p(t'')$ in the Algorithm~\ref{alg:secondlayer} example).
\end{enumerate}

From the perspective of First-Order Linear Temporal Logic (FO-LTL), stacking multiple temporal reasoning layers enables us to realize FO-LTL formulas such as:
\[ p_1~\textit{\textbf{U}}~p_2~\textit{\textbf{U}}~p_3~\textit{\textbf{U}}~\cdots~\textit{\textbf{U}}~p_k, \]
which is interpreted as $p_1$ holds true until $p_2$ becomes true and $p_2$ holds true from that until $p_3$ becomes true $\cdots$, and
\[ \textit{\textbf{F}} p_1~\textit{\textbf{XF}}~p_2~\textit{\textbf{XF}}~p_3~\textit{\textbf{XF}}~\cdots~\textit{\textbf{XF}}~p_k, \]
which is interpreted as $p_1$ eventually becomes true and after that $p_2$ eventually becomes true and after that $\cdots$, and in addition, formulas with interleaved until and eventually quantifiers. Here, $\textit{\textbf{XF}}$ is a composition of the ne\textit{\textbf{X}}t operator and the \textit{\textbf{F}}inally operator in LTL. Meanwhile, as described so far, \modelplural can only nest object quantification inside temporal quantification, so it can represent
\textbf{\textit{always}} $\exists x.~q_1(x) \wedge q_2(x)$, but not $\exists x.~\textbf{\textit{always}}\; q_1(x) \wedge q_2(x)$. This can be solved by interleaving relational and temporal reasoning layers.

It is important to notice that, by using object and temporal pooling operations together with trainable neural networks to realize logic formulas with object and temporal quantifiers, the idea itself generalizes to a broader set of FO-LTL formulas. We design \modelplural to model only a subset of FO-LTL formulas, because they can be computed efficiently (with only $O(T^2)$ space) and they are expressive enough for the type of data we are trying to model.

\section{Experiments}

We compare our model with other approaches to object-centric temporal event detection in this section,
and include an application of \modelplural to concept learning over robot object-manipulation trajectories in the supplementary material.
The setups and metrics focus on data efficiency and generalization.

\mysubsection{Baseline Approaches}
We compare \modelplural against five baselines. The first two are spatial-temporal graph convolutional neural networks \citeb{STGAN;}{yan2018spatial} and its variant STGCN-MAX, which models entity relationships with graph neural networks and models temporal structure with temporal-domain convolutions. The third is STGCN-LSTM, which uses STGCN layers for entity relations but LSTM~\cite{hochreiter1997long} for temporal structures. The last two baselines are based on space-time graphs: Space-Time Graph~\cite{wang2018videos} and Non-Local networks~\cite{wang2018non}. We provide details about our implementation and how we choose the model configurations in the supplementary material.

\mysubsection{Trajectory-Based Soccer Event Detection}

We start our evaluation with an event-detection task in soccer games. The task is to recognize the action performed by a specific player at specific time step in a soccer game trajectory.

\begin{table}[!tp]
\vspace{0pt}
    \centering\small
    \setlength{\tabcolsep}{10.5pt}
    \begin{tabular}{lccc}
    \toprule
        Model & Reg. & Few-Shot  & Full  \\ \midrule
        STGCN & 73.2$_{\pm 1.6}$ & 26.0$_{\pm 5.7}$ & 62.8$_{\pm 0.6}$ \\
        STGCN-MAX & 73.6$_{\pm 1.5}$ & 28.6$_{\pm 5.0}$ & 63.6$_{\pm 0.7}$ \\
        STGCN-LSTM & 72.7$_{\pm 1.4}$ & 23.8$_{\pm 5.9}$ & 61.9$_{\pm 0.6}$ \\
        Space-Time & 74.8$_{\pm 1.5}$ & 31.7$_{\pm 6.1}$ & 65.2$_{\pm 0.6}$ \\
        Non-Local & 76.5$_{\pm 2.4}$ & 45.0$_{\pm 6.3}$ & 69.5$_{\pm 2.4}$ \\
        \model (ours) & \textbf{87.7$_{\pm 1.3}$} & \textbf{52.2$_{\pm 6.3}$} & \textbf{79.8$_{\pm 0.8}$} \\
    \bottomrule
    \end{tabular}
    \vspace{-5pt}
        \captionof{table}{Results on the soccer event dataset. Different columns correspond to different action sets (the regular, few-shot, and full action sets). The performance is measured by per-action (macro) accuracy, averaged over nine few-shot splits. The ${\pm}$ values indicate standard errors. \model significantly outperforms all baseline methods on the few-shot action set.}
        \label{tab:gfootball}
\end{table}

\newmyparagraph{Dataset and setup.}
We collect training and evaluation datasets based on the gfootball simulator\footnote{\url{https://research-football.dev/}}, which provides a physics-based 3D football simulation. It also provides AI agents that can be used to generate random plays.
The simulator provides the 3D coordinates of the ball and the players as well as the action each player is performing at each time step. There are in total 13 actions defined in the simulator, including \textit{movement, ball\_control, trap, short\_pass, long\_pass, high\_pass, header, shot, deflect, catch, interfere, trip} and \textit{sliding}. 
We exclude \textit{header} and \textit{catch} actions, as they never appear in AI games. We also exclude \textit{ball\_control} and \textit{movement}, since they just mean the agent is moving (with or without the ball). Thus, in total, we have nine action categories. We run the simulator with AI-controlled players to generate plays, and formulate the task as classifying the action (9-way classification) of a specific player at a specific time step given a temporal context (25 frames). For each action, we have generated 5,000 videos, expect for \textit{sliding}, for which we generated 4,000 videos because it is rare in the AI games. Among the generated examples, 62\% (2,462 or 3,077) are used for training, 15\% are used for validation, and 23\% are used for testing. Each trajectory is an 8-fps replay clip that contains 17 frames (about two seconds). There is a single ``target'' player in each trajectory. The action label of the trajectory is the action performed by this target player at frame \#9. We randomly split all actions into two categories: seven ``regular'' actions, for which all game plays are available, and two ``few-shot'' actions, for which only 50 clips are available during training. 
\begin{table*}
    \centering\small
    \begin{tabular}{lccccccc}
    \toprule
        Model   & 3v3 & 4v4 & 6v6 & 6v6 \scriptsize{(Time Warp)} & 8v8 & 11v11  \\ \midrule
STGCN &\cellcolor{yellow!25}{\mycellc{40.7$_{\pm 1.0}$~~{\scriptsize (-40.4\%)}}}  & \cellcolor{yellow!25}{\mycellc{63.2$_{\pm 4.9}$~~{\scriptsize (-7.4\%)}}}  & {\mycellc{68.2$_{\pm 2.8}$}} & \cellcolor{blue!5}{\mycellc{52.8$_{\pm 7.0}$~~{\scriptsize (-22.6\%)}}}  & \cellcolor{yellow!25}{\mycellc{55.4$_{\pm 3.3}$~~{\scriptsize (-18.8\%)}}}  & \cellcolor{yellow!25}{\mycellc{44.4$_{\pm 2.1}$~~{\scriptsize (-34.9\%)}}} \\
STGCN-MAX & \cellcolor{yellow!25}{\mycellc{47.4$_{\pm 3.2}$~~{\scriptsize (-33.7\%)}}}  & \cellcolor{yellow!25}{\mycellc{68.8$_{\pm 2.0}$~~{\scriptsize (-3.8\%)}}}  & {\mycellc{71.5$_{\pm 1.9}$}} & \cellcolor{blue!5}{\mycellc{56.5$_{\pm 4.5}$~~{\scriptsize (-21.0\%)}}}  & \cellcolor{yellow!25}{\mycellc{59.1$_{\pm 0.7}$~~{\scriptsize (-17.3\%)}}}  & \cellcolor{yellow!25}{\mycellc{45.6$_{\pm 2.5}$~~{\scriptsize (-36.2\%)}}} \\
STGCN-LSTM & \cellcolor{yellow!25}{\mycellc{39.7$_{\pm 1.1}$~~{\scriptsize (-43.1\%)}}}  & \cellcolor{yellow!25}{\mycellc{60.4$_{\pm 0.2}$~~{\scriptsize (-13.5\%)}}}  & {\mycellc{69.8$_{\pm 0.1}$}} & \cellcolor{blue!5}{\mycellc{30.6$_{\pm 0.6}$~~{\scriptsize (-56.1\%)}}}  & \cellcolor{yellow!25}{\mycellc{55.8$_{\pm 2.0}$~~{\scriptsize (-20.0\%)}}}  & \cellcolor{yellow!25}{\mycellc{44.1$_{\pm 0.7}$~~{\scriptsize (-36.8\%)}}} \\
Space-Time & \cellcolor{yellow!25}{\mycellc{29.0$_{\pm 1.6}$~~{\scriptsize (-60.4\%)}}}  & \cellcolor{yellow!25}{\mycellc{53.5$_{\pm 3.2}$~~{\scriptsize (-27.0\%)}}}  & {\mycellc{73.3$_{\pm 0.3}$}}  & \cellcolor{blue!5}{\mycellc{70.7$_{\pm 0.3}$~~{\scriptsize (-3.5\%)}}} & \cellcolor{yellow!25}{\mycellc{33.9$_{\pm 2.8}$~~{\scriptsize (-53.7\%)}}}  & \cellcolor{yellow!25}{\mycellc{15.2$_{\pm 1.8}$~~{\scriptsize (-79.3\%)}}} \\
Non-Local & \cellcolor{yellow!25}{\mycellc{45.9$_{\pm 5.1}$~~{\scriptsize (-41.2\%)}}}  & \cellcolor{yellow!25}{\mycellc{70.7$_{\pm 5.3}$~~{\scriptsize (-9.5\%)}}}  & {\mycellc{78.1$_{\pm 5.8}$}} & \cellcolor{blue!5}{\mycellc{77.7$_{\pm 5.0}$~~{\scriptsize (-0.5\%)}}}  & \cellcolor{yellow!25}{\mycellc{58.5$_{\pm 10.8}$~~{\scriptsize (-25.1\%)}}}  & \cellcolor{yellow!25}{\mycellc{41.8$_{\pm 13.6}$~~{\scriptsize (-46.5\%)}}} \\
\model & \cellcolor{yellow!25}{\mycellc{\bf 77.4$_{\pm 3.5}$~~{\scriptsize (-12.4\%)}}}  & \cellcolor{yellow!25}{\mycellc{\bf 88.3$_{\pm 0.7}$~~{\scriptsize (-0.0\%)}}}  & {\mycellc{\bf 88.4$_{\pm 0.6}$}} & \cellcolor{blue!5}{\mycellc{\bf 86.9$_{\pm 0.4}$~~{\scriptsize (-1.7\%)}}}  & \cellcolor{yellow!25}{\mycellc{\bf 81.3$_{\pm 1.7}$~~{\scriptsize (-8.0\%)}}}  & \cellcolor{yellow!25}{\mycellc{\bf 77.1$_{\pm 1.7}$~~{\scriptsize (-12.8\%)}}} \\
    \bottomrule
    \end{tabular}
    \vspace{-5pt}
    \caption{Results on generalization to scenarios with more agents and temporally warped trajectories on the soccer event dataset.
    All models are trained only on 6v6 games. The standard errors indicated by the $\pm$ signs are computed with three random seeds.
    }
    \label{tab:gfootball-generalization}
\end{table*}

\newmyparagraph{Input features.}
Each trajectory is represented with 7 time-varying unary predicates, including the 3D coordinate of each player and the ball and four extra predicates defining the type of each entity $x$: $\mathrm{IsBall}(x)$, $\mathrm{IsTargetPlayer}(x)$, $\mathrm{SameTeam}(x)$, $\mathrm{OpponentTeam}(x)$, where $\mathrm{SameTeam}(x)$ and $\mathrm{OpponentTeam}(x)$ indicates whether $x$ is of the same team as the target player. We also add a temporal indicator function which is a Gaussian function centered at frame of interest with variance $\sigma^2=25$.

\newmyparagraph{Results.}
\tbl{tab:gfootball} shows the result. Our model significantly outperforms all the baselines in all three action settings,  suggesting that our model is able to discover a set of useful features at both input and intermediate levels and use them to compose new action classifiers from only a few examples.

\newmyparagraph{Generalization to more players.} Due to their object-centric design, \modelplural can generalize to soccer games with a varying number of agents. After training on 6v6 soccer games (\ie, 6 players on each team), we evaluate the performance of different models on games with different numbers of players: 3v3, 4v4, 8v8, and 11v11.
For each action we have generated, on average, 1,500 examples for testing. \tbl{tab:gfootball-generalization} summarizes the result and the full results are provided in the supplementary material.
Comparing the columns highlighted in yellow, we notice a significant performance drop for all baselines while \model performs the best.
By visualizing data and predictions, we found that misclassifications of instances of {\it shot} as {\it short pass} contribute most to the performance degradation of our model when we have more players. Specifically, the recall of {\it shot} drops from 97\% to 60\%. In soccer plays with many agents, a shot is usually unsuccessful and a player from another team steals the ball in the end. In such scenarios, \model tends to misclassify such trajectories as a {\it short pass}. Ideally, this issue should be addressed by understanding actions based on agents' goals instead of the actual outcome~\cite{intille2001recognizing}. We leave this extension as a future direction.

\newmyparagraph{Generalization to temporally warped trajectories.} Another crucial property of \modelplural is to recognize actions based on their sequential order in the input trajectory, instead of binding features to specific time steps in the trajectory. To show this, we test the performance of different models on time warped trajectories. Each test trajectory has a length of 25, and each trajectory is labeled by the action performed by the target player at any time step between the 6-th and 19-th frame. We ensure that the target player performs only one action during the entire input trajectory. Thus, the label is unambiguous.
The results are shown in \tbl{tab:gfootball-generalization}. Specifically, our test set consists of 25-frame trajectories, and the action may occur at anytime between the 6th and the 19th frame. By comparing rows with and without time warping, we notice a $60\%$ performance drop for STGCN, STGCN-MAX, and STGCN-LSTM. In contrast, \modelplural still have reasonable performance. Note that Space-Time and Non-Local model have almost no performance drop against time warping because they are completely agnostic to temporal ordering.

\mysubsection{Extension to Real-World Datasets}
The proposed \model can also be extended to other real-world datasets. %
These examples further illustrate the robustness of TOQ-Net to temporal variations in activities.

\subsubsection{Toyota Smarthome}
Toyota Smarthome~\cite{das2019toyota} is a dataset that contains videos of humans performing everyday activities such as ``walk'', ``take pills'', and ``use laptop''. It also comes with 3D-skeleton detections. There are around 16.1k videos in the dataset, and 19 activity classes. The videos’ length varies between a few seconds to 3 minutes. We subsample 30 frames for each video. We split frames into training (9.9k), validation (2.5k), and testing (3.6k). We treat human joints as entities. The input is then the position of joints, the velocity of joints, limb lengths, and joint angles. We evaluated our model and STGCN on a 19-way classification task. We also test model performance on time-warped sequences by accelerating the trajectories by two times.

Our model achieves a comparable accuracy to STGCN on the standard classification task: (42.0\% vs. 43.0\%). Importantly, on the generalization test to time-warped sequences, our model has only a 0.8\% performance drop (41.2\%), while STGCN drops 10.7\% (32.3\%). This indicates that the temporal structures learned by \model improve model generalization to varying time courses.

\subsubsection{Volleyball Activity}
The volleyball dataset~\cite{msibrahi2016deepactivity} contains 4830 video clips collected from 55 youtube volleyball videos. They are labeled with 8 group activities (e.g. “left spike” and “right pass”). Each video contains 20 frames with the labeled group activity performed at the 10-th frame. The dataset also includes annotations for players, including the bounding box, the indicator of whether the player is involved in the group activity, and the individual action such as ``setting'', ``digging'', and ``spiking''. We use the manual annotations (processed by an MLP) as the input features. We train models to classify the video into one of the eight group activities, following the original split, i.e., 24, 15, and 16 of 55 videos are used for training, validation, and testing.

On the standard classification task, TOQ-Net achieves a comparable performance with STGCN (73.3\% vs. 73.6\%). When we perform time warping on the input sequences, STGCN’s performance drops by more than 25.0\% (39.5\% on temporally shifted trajectories and 48.6\% on 2$\times$ quick motion trajectories), while our model drops only 3\% (70.3\% on temporally shifted trajectories and 70.7\% on quick motion trajectories). This again shows the generalization ability of TOQ-Net \wrt varying time courses, and the robustness of learned temporal structures.

%
\section{Related Work}

\newmyparagraph{Action concept representations and learning.}
First-order and linear temporal logics~\citeb{LTL;}{pnueli1977temporal}
have been used for analyzing sporting events~\cite{intille1999framework,intille2001recognizing} and activities of daily living~\cite{tran2008event,brendel2011probabilistic} in logic-based reasoning frameworks. However, these frameworks require extra knowledge to annotate relationships between low-level, primitive actions and complex ones, or performing search in a large combinatorial space for candidate temporal logic rules~\cite{penning2011nsca,lamb2007connectionist}. By contrast, \modelplural enable end-to-end learning of complex action descriptions from sensory input with only high-level action-class labels.

\newmyparagraph{Temporal and relational reasoning.} 
This paper is also related to work on using data-driven models for modeling relational and temporal structure, such as LTL ~\cite{neider2018learning,camacho2018finite,chou2020explaining}, Logical Neural Networks~\cite{riegel2020logical}, ADL description languages~\cite{intille1999framework}, and hidden Markov models~\cite{tang2012learning}. These models need human-annotated action descriptions and symbolic state variables (\eg, {\it pick up} $x$ means a state transition from not {\it holding} $x$ to {\it holding} $x$), and dedicated inference algorithms such as graph structure learning. In contrast, \modelplural have an end-to-end design, and can be integrated with other neural networks.
People have also used structural representations to model object-centric temporal concepts with graph neural networks~\citeb{GNNs;}{yan2018spatial}, recurrent neural networks~\citeb{RNNs;}{msibrahi2016deepactivity}, and integrated GNN-RNN architectures \cite{deng2016sim,qi2018stag}.
\modelplural use a similar relational representation, but different models for temporal structures.
\section{Conclusion and Discussion}
The design of \modelplural suggests multiple research directions. For example, the generalization of the acquired action concepts to novel object kinds, such as from {\it opening fridges} to {\it opening bottles}, needs further exploration. Meanwhile, \modelplural are based on physical properties, \eg 6D poses. Incorporating representations of mental variables such as goals, intentions, and beliefs can aid in action and event recognition~\cite{baker2017rational,zacks2001perceiving,vallacher1987people}.

In summary, we have presented \modelplural, a neuro-symbolic architecture for learning to describe complex state sequences with quantification over both entities and time.  \modelplural use tensors to represent the time-varying properties and relations of different entities, and use tensor pooling operations over different dimensions to realize temporal and object quantifiers. 
\modelplural generalize well to scenarios with varying numbers of entities and time courses.

\section*{Acknowledgements} We thank Tom\'as Lozano-P\'erez for helpful discussions and suggestions. We gratefully acknowledge support from the Center for Brains, Minds and Machines (NSF STC award CCF1231216), ONR MURI N00014-16-1-2007, NSF grant 1723381, AFOSR grant FA9550-17-1-0165, ONR grant N00014-18-1-2847, the Honda Research Institute, MIT-IBM Watson Lab, SUTD Temasek Laboratories, the Samsung Global Research Outreach (GRO) Program, and the Stanford Institute for Human-Centered AI (HAI). Any opinions, findings, and conclusions or recommendations expressed in this material are those of the authors and do not necessarily reflect the views of our sponsors.

\clearpage

{
\bibliographystyle{named}
\bibliography{actiongrounding}
}

\end{document}


\maketitle

We start from inrtoducing a new dataset  we collected for model evaluation, namely the RLBench dataset (\sectapp{app:rlbench}). We then provide implementation details of different models in \sectapp{app:implementation}. Finally, in \sectapp{app:additional}, we present additional ablation studies and visualizations.

\appendix

\section{Manipulation Concept Learning from 6D Poses}
\label{app:rlbench}

Structural action representations can also be usefully applied to other domains. Here we show the result in a robotic environment, where the goal is to classify the action performed by a robotic arm.

\newmyparagraph{Dataset and setup.}
We generated a dataset based on the RLBench simulator~\cite{james2019rlbench}, which contains a set of robotic object-manipulation actions in a tabletop environment. We use 24 actions from the dataset, including \textit{CloseBox, CloseDrawer, CloseFridge, CloseGrill, CloseJar, CloseLaptopLid, CloseMicrowave, GetIceFromFridge, OpenBox, OpenFridge, OpenMicrowave, OpenWineBottle, PickUpCup, PressSwitch, PushButtons, PutGroceriesInCupboard, PutItemInDrawer, PutRubbishInBin, PutTrayInOven, ScoopWithSpatula, SetTheTable, SlideCabinetOpenAndPlaceCups, TakePlateOffColoredDishRack,} and \textit{ TakeToiletRollOffStand}. 
We randomly initialize the position and orientation of different objects in the scene and use the built-in motion planner to generate robot arm trajectories. Depending on the task and the initial poses of different objects, trajectories may have different lengths, varying from 30 to 1,000. Most of the trajectories have $\sim$50 frames.

For each action category, we have generated 100 trajectories. Among the generated examples, 60\% of the examples are used for training, 15\% are used for validation, and the other 25\% are used for testing.

\paragraph{Input.}
Each trajectory contains the poses of different objects, at each time step. Each object except for the robot arm is represented as 13 unary predicates, including the 3D position, the 3D orientation represented as quaternions (4D vector), and the 3D bounding box (axis-aligned, 6D vector). The robot arm is represented as 8 unary predicates, including the 3D position, the 3D orientation, and a binary predicate that indicates whether the gripper is open or closed. Note that deformable objects are split into several pieces so that the learner can infer their state (open or closed). For example, the lid of the box is a separate object. We also input the 6D pose and the state of the robot gripper. During training, we used on average 65 robot trajectories for regular actions. 

\newmyparagraph{Input features.}
All methods take the raw trajectories as input. In \modelplural, we use an extra single linear layer with sigmoid activation as the input feature extractor.
To evaluate temporal modeling with recurrent neural networks, STGCN-LSTM uses kernel size $1$ in its (STGCN) temporal convolution layers. Ablation studies can be found in \sectapp{app:implementation}.

\begin{table}[ht]
    \begin{minipage}[t]{0.5\textwidth}
    \centering\small
    \setlength{\tabcolsep}{3pt}
    \vspace{2pt}
    \begin{tabular}{llcc}
    \toprule
         Model & Reg. & 1-Shot & Full   
         \\ \midrule
        STGCN & 99.89$_{\pm 0.05}$ & 94.92$_{\pm 1.03}$ & 98.79$_{\pm 0.23}$  
        \\
        STGCN-LSTM & \bf{99.92$_{\pm 0.03}$} & 95.48$_{\pm 1.67}$ & \bf{98.86$_{\pm 0.43}$}  
        \\
        \midrule
        \model & {\bf 99.96$_{\pm 0.02}$} & {\bf 98.04$_{\pm 0.97}$} & {\bf 99.48$_{\pm 0.24}$} 
        \\
    \bottomrule
    \end{tabular}
    \centering
    \caption{\label{tab:rlbench}Few-shot learning on the RLBench dataset, measured by per-action (macro) accuracy and averaged of four 1-shot splits and four random seeds per split. The ${\pm}$ values indicate standard errors. On the right we shows the sampled performance of different models on each individual 1-shot split.}
    \end{minipage}
    \begin{minipage}[t]{0.5\textwidth}
    \vspace{0pt}
    \includegraphics[width=\textwidth]{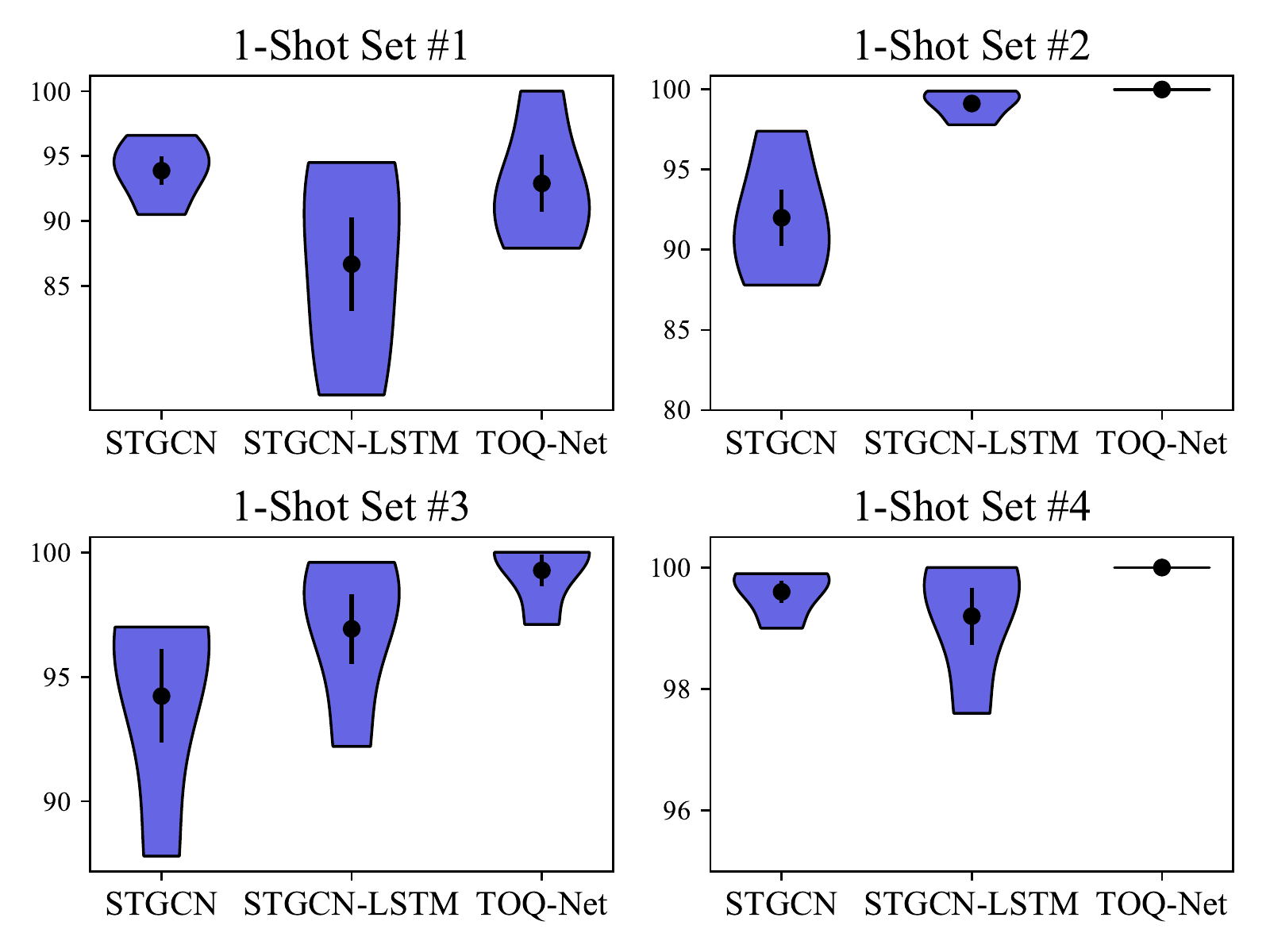}
    \end{minipage}
\end{table} 
\begin{figure}{l}
    \centering
    \vspace{-5pt}
    \includegraphics[width=0.45\textwidth]{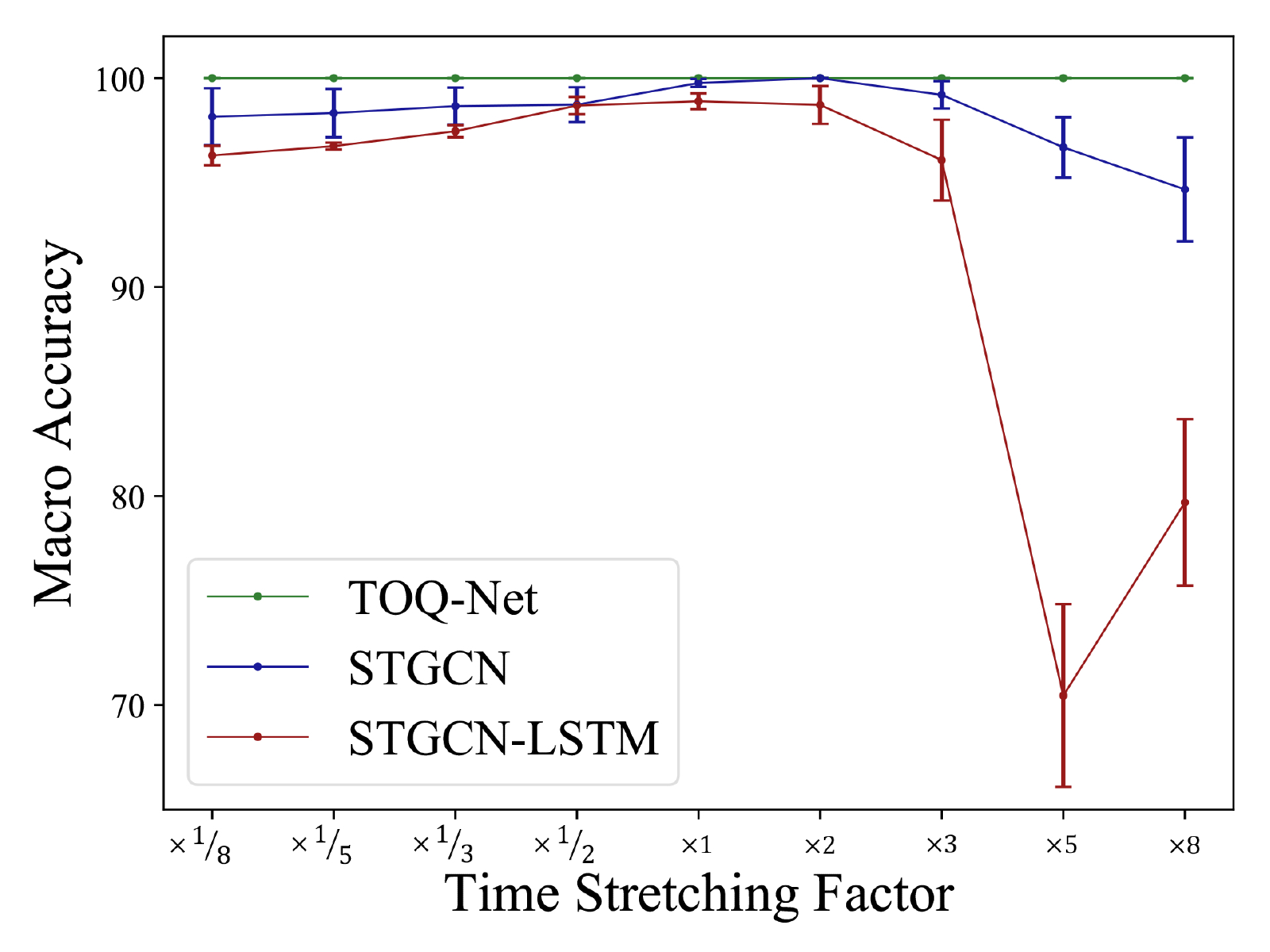}
    \caption{Comparing different models with different time stretching factors on the RLBench dataset.}
    \label{fig:rlbench-slowfast}
    \vspace{-5pt}
\end{figure}
 
\newmyparagraph{Results.}
We start with evaluating different models on the standard 24-way classification task. We summarize results in \fig{fig:rlbench-slowfast}. It also illustrates each model's performance under different time stretching factors of the input trajectory, from $\nicefrac{1}{8}\times$ to $8\times$. Similar to the soccer dataset, the \model outperforms both convolutional and recurrent models across all time stretching factors.

\newmyparagraph{Generalization to new action concepts.}
We also evaluate how well different models can generalize the learned knowledge about opening and closing other objects to novel objects. Specifically, we hold out 50\% of all {\it Open X} actions and 50\% of the {\it Close X} actions to form the 1-shot action learning set. For example, the learner may only see a single instance of the opening of a box during training. Actions with more examples are ``regular.''  All models are tested on a 3-way classification task: {\it Open-X}, {\it Close-X}, and {\it Other}. Results are summarized in \tbl{tab:rlbench}.
There are several held-out sets that are noticeably more ``difficult'' than others , such as the Set {\#1} in \tbl{tab:rlbench}. Our visualization shows that actions {\it close jar} and {\it close drawer} are harder  generalization targets, compared to other actions. This is possibly because the motion of {\it close jar} involves some rotation, and the motion of {\it close drawer} looks more like pushing than other {\it close-X} actions.

\section{Implementation Details}
\label{app:implementation}

In this section, we present the implementation detail of our model, the \modelplural, and five baselines (STGCN, STGCN-LSTM, STGCN-MAX, Space-Time and Non-local), including the model architecture, input features, and the training methods.

\subsection{\modelplural}

In the soccer event detection task, we use three object quantification layers and three temporal quantification layers. Each object quantification layer has a hidden dimension of 16 (\ie, all nullary, unary, and binary tensors have the same hidden dimension of 16). Each temporal quantification layer has a hidden dimension of 48.

In the manipulation concept learning task, we use three object quantification layers and three temporal quantification layers. Each object quantification layer has a hidden dimension of 48. Each temporal quantification layer has a hidden dimension of 144.

\paragraph{Input feature extractor.}

We use the following physics-based input feature extractor for soccer event detection. Given the trajectories of all entities, including all players and the ball, we first compute the following physical properties: ground speed (\ie, the velocity on the $xy$-plane), vertical speed (\ie, the velocity along the $z$ direction), and acceleration. We also compute the distance between each pair of entities.

After extracting the physical properties, we use the following feature templates to generate the input features. For each physical property, $X$, we first normalize $X$ across all training trajectories, and then we create $c=5$ features:
\[
\sigma \left(\frac{X-\theta_i}{\beta}\right),
\]
where $\sigma$ is the sigmoid function, $\beta$ is a scalar hyperparameter, $\theta_i, i=1,2,3,4,5$ are trainable parameters. These feature templates can be interpreted as differentiable implementations of $\mathbf{1}[X>\theta_i]$, where $\mathbf{1}[\cdot]$ is the indicator function.
During training, we make $\beta$ decay exponentially from 1 to 0.001.

In the manipulation concept learning task, we use a single fully-connected layer with sigmoid activation as the feature extractor. The hidden dimension of the layer is 64.

\subsection{STGCN}
\label{app:baseline}

We use the same architecture as Yan~\etal~\cite{yan2018spatial}. \tbl{tab:stgcn-architecture} summarizes the hidden dimensions and the kernel sizes.

The STGCN model's output is a tensor of size $(T/4) \times N \times 256$, where $T$ is the length of the input trajectory, and $N$ is the number of entities. Following Yan~\etal~\cite{yan2018spatial}, we perform an average pooling over the temporal and the entity dimension and get a feature vector of size 256. We apply a softmax layer on top of the latent feature to classify the action.

\paragraph{Input feature extractor.}

In the soccer event detection task, we do not input the distance between each pair of entities as we do for the \modelplural, as the STGCN model does not support binary predicate inputs (see STGCN-2D for an ablation study).
In the manipulation concept learning task, STGCN uses the same input format as the \modelplural. Specifically, we represent the state of each entity with 13 unary predicates, including 3D position, 3D orientation, and bounding boxes.

\subsection{STGCN-MAX}

The STGCN-MAX model is a variation of STGCN. During graph propogation, for each node, instead of taking average of all propagated information from all neighbors, we take their maximum value over all feature dimensions, and update the hidden state of each node with this. This baseline replicates a very similar propagation rule as our relational reasoning layer. However, it still uses temporal convolutions to model temporal structures.

\subsection{STGCN-2D}
The original STGCN does not support input features of higher dimensions, so we extend STGCN to STGCN-2D to add binary inputs such as the distance. This extension slightly improves the model performance on the standard 9-way classification test. It also helps in the few-shot setting. Specifically, on the few-shot learning setting:

\begin{table}[!t]
\centering
\begin{tabular}{cccc}
\toprule
     & Reg. & New & All  \\
   \midrule
   \model  & \textbf{87.7} & \textbf{52.2} &  \textbf{79.8} \\
   STGCN-2D  & 84.1 & 46.9 & 75.8\\
\bottomrule
\end{tabular}
\caption{Ablation study of the STGCN-2D model on the soccer event dataset, evaluated by its few-shot learning performance.}
\end{table}

Meanwhile, adding extra binary inputs does not improve the generalization of the network to games with a different number of agents or to time-warped sequences. This following table extends the Table 2 in the main text. Specifically, we train STGCN-2D on 6v6 soccer games and test it on scenarios with a different number of agents (3v3 and 11v11) and also temporally warped trajectories. Our model shows a significant advantage.

\begin{table}[!t]

\centering
\begin{tabular}{ccccc}
\toprule
  & 9-way & 3v3 & 11v11 & 6v6(Time warp) \\
  \midrule
  TOQ-Net &	\textbf{88.4} & \textbf{77.4} & \textbf{77.1} & \textbf{86.9} \\
  STGCN-2D & 84.5 & 17.5 & 16.6 & 39.7\\
  \bottomrule
\end{tabular}
\caption{Ablation study of the STGCN-2D model on the soccer event dataset, evaluated by its generalization to different number of players and time-warped trajectories.}
\end{table}

\subsection{STGCN-LSTM}

The STGCN-LSTM model first encodes the input trajectory with an STGCN module. The output of the STGCN module is a tensor of shape $T \times N \times 256$, where $T$ is the length of the input trajectory, and $N$ is the number of entities. We perform an average pooling over the entity dimension and get a tensor of shape $T \times 256$. Next, we encode this feature with a 2-layer bidirectional LSTM module. The hidden dimension for the LSTM module is 256. We use a single softmax layer on top of the LSTM module to classify the action.

STGCN-LSTM uses the same architecture as STGCN for both tasks, except for the kernel size and the stride, because our goal is to evaluate the performance of recurrent neural networks (the LSTM model) on modeling temporal structures. In the soccer event detection task, we use kernel size 3 and stride 1 so that the STGCN module will have the representational power to encode local physical properties. In the manipulation concept learning task, we use a temporal kernel size of 1 and stride 1. Meanwhile, \tbl{tab:rlbench-fewshot-full} shows the ablation study of different kernel sizes, evaluated by the few-shot learning performance on the RLBench dataset. Different kernel sizes (1 and 3) have a similar performance.

\subsection{Space-Time Region Graph Networks}

We use the same architecture for each Space-Time layer as Wang \etal proposed in Space-Time Graph~\cite{wang2018videos}. \tbl{tab:strg-architecture} summarises the detailed parameters of all layers.

\subsection{Non-Local Neural Networks}
We use the same architecture for each Non-local layer as Wang \etal proposed in Non-Local networks~\cite{wang2018non}. The original non-local network works on pixels. Here instead, we apply it to the space-time graph constructed in the same way as \citeauthor{wang2018videos}~\citeyear{wang2018videos}. The major difference between the Space-Time Graph and the Non-local Network is that Space-Time Graph uses graph convolutions while Non-local uses transfomer-like attentions over all nodes in the graph. \tbl{tab:nonlocal-architecture} summarises the detailed parameters of all layers.

\begin{table}[t]
    \centering
    \small
    \setlength{\tabcolsep}{5pt}
    \begin{tabular}{llcc}
    \toprule
         Model & Reg. & 1-Shot & Full   
         \\ \midrule
        STGCN & 99.89$_{\pm 0.05}$ & 94.92$_{\pm 1.03}$ & 98.79$_{\pm 0.23}$  
        \\
        \mycell{STGCN-LSTM \\ (kernel size 3)} & 99.89$_{\pm 0.05}$ & 96.16$_{\pm 1.39}$ & 99.04$_{\pm 0.31}$
        \\
        \mycell{STGCN-LSTM \\ (kernel size 1)} & 99.92$_{\pm 0.03}$ & 95.48$_{\pm 1.67}$ & 98.86$_{\pm 0.43}$ 
        \\
        \midrule
        \model & {\bf 99.96$_{\pm 0.02}$} & {\bf 98.04$_{\pm 0.97}$} & {\bf 99.48$_{\pm 0.24}$} 
        \\
    \bottomrule
    \end{tabular}
    \vspace{5pt}
    \caption{Ablation study of different kernel sizes for the STGCN-LSTM model on the RLBench dataset, measured by per-action (macro) accuracy and averaged of four 1-shot splits and four random seeds per split. The ${\pm}$ values indicate standard errors.}
    \label{tab:rlbench-fewshot-full}
\end{table} 
\paragraph{Soccer event detection.}
We use the identical training strategy for our model and both baselines. Each training batch contains 128 examples, which are sampled as following: we first uniformly sample an action category, and then uniformly sample a trajectory labeled as this action category.
 
In all the soccer event tasks, we optimized our model using Adam~\cite{kingma2014adam}, with the learning rate initialized to $\eta=0.002$. The learning rate is fixed for the first 50 epochs. After that, we decrease the learning rate $\eta$ by a factor of 0.9 if the best validation loss has not been updated during the last 6 epochs.

All models are trained for 200 epochs. In the task of generalization to different input scales, we first train different models on the original dataset for 200 epochs ($\sim$40,000 iterations) and finetune them on the new dataset with few samples for 10,000 iterations.

\paragraph{Manipulation concept learning.}
We optimized different models using Adam with the initial learning rate $\eta=0.003$. We applied the same learning rate schedule as in the soccer event detection task. The batch size for the manipulation concept learning task is 32 for models with STGCN backbone and is 4 for \modelplural. All models are trained for 150 epochs.

\begin{table*}[t]
    \centering
    \begin{tabular}{c|cccccc}
    \toprule
        Model & Input Dim. & Output Dim. & Kernel Size & Stride & Dropout & Residual  \\ \midrule
        \multirow{10}{*}{STGCN}
        & - & 64 & 7 & 1 & 0 & False \\
        & 64 & 64 & 7 & 1 & 0.5 & True \\
        & 64 & 64 & 7 & 1 & 0.5 & True \\
        & 64 & 64 & 7 & 1 & 0.5 & True \\
        & 64 & 128 & 7 & 2 & 0.5 & True \\
        & 128 & 128 & 7 & 1 & 0.5 & True \\
        & 128 & 128 & 7 & 1 & 0.5 & True \\
        & 128 & 256 & 7 & 2 & 0.5 & True \\
        & 256 & 256 & 7 & 1 & 0.5 & True \\
        & 256 & 256 & 7 & 1 & 0 & True \\
        \midrule
        \multirow{6}{*}{STGCN$_S$}
        & - & 16 & 7 & 1 & 0 & False \\
        & 16 & 16 & 7 & 1 & 0.5 & True \\
        & 16 & 32 & 7 & 2 & 0.5 & True \\
        & 32 & 32 & 7 & 1 & 0.5 & True \\
        & 32 & 64 & 7 & 2 & 0.5 & True \\
        & 64 & 128 & 7 & 1 & 0 & True \\
        \midrule
        \multirow{3}{*}{STGCN$_T$}
        & - & 8 & 7 & 1 & 0 & False \\
        & 8 & 12 & 7 & 2 & 0.5 & True \\
        & 12 & 64 & 7 & 1 & 0 & True \\
    \bottomrule
    \end{tabular}
    \vspace{5pt}
    \caption{The STGCN architecture used in the paper.}
    \label{tab:stgcn-architecture}
\end{table*}

\begin{table*}[t]
    \centering
    \begin{tabular}{c|cccccc}
    \toprule
        Model & Input Dim. & Output Dim.  & Stride & Residual  \\ \midrule
        \multirow{10}{*}{Space-Time}
        & - & 64 & 1 & False \\
        & 64 & 64 & 1 & True \\
        & 64 & 64 & 1 & True \\
        & 64 & 64 & 1 & True \\
        & 64 & 128 & 2 & True \\
        & 128 & 128 & 1 & True \\
        & 128 & 128 & 1 & True \\
        & 128 & 256 & 2 & True \\
        & 256 & 256 & 1 & True \\
        & 256 & 256 & 1 & True \\
        \midrule
        \multirow{5}{*}{Space-Time$_S$}
        & - & 16 & 1 & False \\
        & 16 & 16 & 1 & True \\
        & 16 & 32 & 2 & True \\
        & 32 & 64 & 2 & True \\
        & 64 & 128 & 1 & True \\
        \midrule
        \multirow{3}{*}{Space-Time$_T$}
        & - & 32 & 1 & False \\
        & 32 & 64 & 2 & True \\
        & 64 & 64 & 2 & True \\
    \bottomrule
    \end{tabular}
    \vspace{5pt}
    \caption{The Space-Time Region Graph architecture used in the paper.}
    \label{tab:strg-architecture}
\end{table*}

\begin{table*}[t]
    \centering
    \begin{tabular}{c|cccccc}
    \toprule
        Model & Input Dim. & Output Dim.  & Stride & Residual  \\ \midrule
        \multirow{10}{*}{Non-Local}
        & - & 64 & 1 & False \\
        & 64 & 64 & 1 & True \\
        & 64 & 64 & 1 & True \\
        & 64 & 64 & 1 & True \\
        & 64 & 128 & 2 & True \\
        & 128 & 128 & 1 & True \\
        & 128 & 128 & 1 & True \\
        & 128 & 256 & 2 & True \\
        & 256 & 256 & 1 & True \\
        & 256 & 256 & 1 & True \\
        \midrule
        \multirow{4}{*}{Non-Local$_S$}
        & - & 32 & 1 & False \\
        & 32 & 32 & 2 & True \\
        & 32 & 64 & 2 & True \\
        & 64 & 128 & 1 & True \\
        \midrule
        \multirow{3}{*}{Non-Local$_T$}
        & - & 32 & 1 & False \\
        & 32 & 32 & 2 & True \\
        & 32 & 64 & 2 & True \\
    \bottomrule
    \end{tabular}
    \caption{The Non-local Neural Networks architecture used in the paper.}
    \vspace{-5pt}
    \label{tab:nonlocal-architecture}
\end{table*}

\section{Additional Ablation Study}
\label{app:additional}

\subsection{Network Architecture}
\begin{table*}[!tp]
    \centering
    \setlength{\tabcolsep}{8pt}
    \vspace{5pt}
    \begin{tabular}{lcccc}
    \toprule
        Model  & \#params & Reg. & Few-Shot  & Full  \\ \midrule
        STGCN   & 3.42M & 72.0$_{\pm 1.5}$ & 22.6$_{\pm 5.2}$ & 61.0$_{\pm 0.8}$ \\
        STGCN$_S$   & 263K & 73.0$_{\pm 1.5}$ & 22.5$_{\pm 4.9}$ & 61.7$_{\pm 0.5}$ \\
        \textbf{STGCN$_T$}   & 34K & 73.2$_{\pm 1.6}$ & 26.0$_{\pm 5.7}$ & \textbf{62.8$_{\pm 0.6}$} \\
        \midrule
        STGCN-LSTM   & 2.08M & 72.5$_{\pm 1.6}$ & 20.2$_{\pm 5.0}$ & 60.9$_{\pm 0.7}$ \\
        STGCN-LSTM$_S$   & 233K & 72.1$_{\pm 1.7}$ & 20.5$_{\pm 5.9}$ & 60.6$_{\pm 0.7}$ \\
        \textbf{STGCN-LSTM$_T$}   & 33K & 72.7$_{\pm 1.4}$ & 23.8$_{\pm 5.9}$ & \textbf{61.9$_{\pm 0.6}$} \\
        \midrule
        STGCN-MAX   & 3.42M & 73.5$_{\pm 4.8}$ & 20.2$_{\pm 6.0}$ & 61.7$_{\pm 4.1}$ \\
        \textbf{STGCN-MAX$_S$}   & 263K & 73.6$_{\pm 1.5}$ & 28.6$_{\pm 5.0}$ & \textbf{63.6$_{\pm 0.7}$} \\
        STGCN-MAX$_T$   & 34K & 74.5$_{\pm 1.2}$ & 25.4$_{\pm 6.1}$ & 63.6$_{\pm 0.7}$ \\
        \midrule
        Space-Time   & 263K & 71.3$_{\pm 1.4}$ & 26.9$_{\pm 7.1}$ & 61.4$_{\pm 1.2}$ \\
        Space-Time$_S$   & 24K & 74.7$_{\pm 1.4}$ & 30.7$_{\pm 6.8}$ & 64.9$_{\pm 0.6}$ \\
        \textbf{Space-Time$_T$}   & 14K & 74.8$_{\pm 1.5}$ & 31.7$_{\pm 6.1}$ & \textbf{65.2$_{\pm 0.6}$}  \\
        \midrule
        Non-Local   & 1.23M & 74.5$_{\pm 4.2}$ & 44.3$_{\pm 7.0}$ & 67.8$_{\pm 3.7}$ \\
        \textbf{Non-Local$_S$}   & 108K & 76.5$_{\pm 2.4}$ & 45.0$_{\pm 6.3}$ & \textbf{69.5$_{\pm 2.4}$} \\
        Non-Local$_T$   & 32K & 75.6$_{\pm 1.2}$ & 44.8$_{\pm 6.4}$ & 68.8$_{\pm 1.0}$ \\
        \midrule
        \model   & 35K & 87.7$_{\pm 1.3}$ & 52.2$_{\pm 6.3}$ & 79.8$_{\pm 0.8}$ \\
    \bottomrule
    \end{tabular}
    \vspace{5pt}
        \captionof{table}{Results on the soccer event dataset for baselines with different capacities, where model$_S$ and model$_T$ denote the small and the tiny variant for each model. The performance is measured by per-action (macro) accuracy, averaged over nine few-shot splits.  The $\pm$ values indicate standard errors. For each baseline we showed the best performance over three levels of capacities. In fact, performances of most of the models are not affected much by the capacity. We highlight the best-performing variants of all baselines, and use them in all comparisons in the main text.}
        \label{tab:gfootball-capacities}
\end{table*}
 
In general, \model has a smaller number of weights than baselines. We add additional comparisons to models of different \#params. Specifically, model$_S$ is a smaller version of the model and model$_T$ is the tiny version of the model whose \#params are about equal to or smaller than \#params of \model. To obtain smaller models we typically reduce the number of layers and the hidden state dimensions. On the few-shot task of the soccer event dataset, we test all baselines with all combination of features and capacities. \tbl{tab:gfootball-capacities} summarizes the results. In general, the ``small'' variations are the best for all models (except for STGCN-LSTM). Across all experiments presented in the main paper, we use the best architecture in \tbl{tab:gfootball-capacities}.

\subsection{Generalization to different input scales}

\begin{figure*}
\vspace{0pt}
    \centering
    \includegraphics[width=0.6\textwidth]{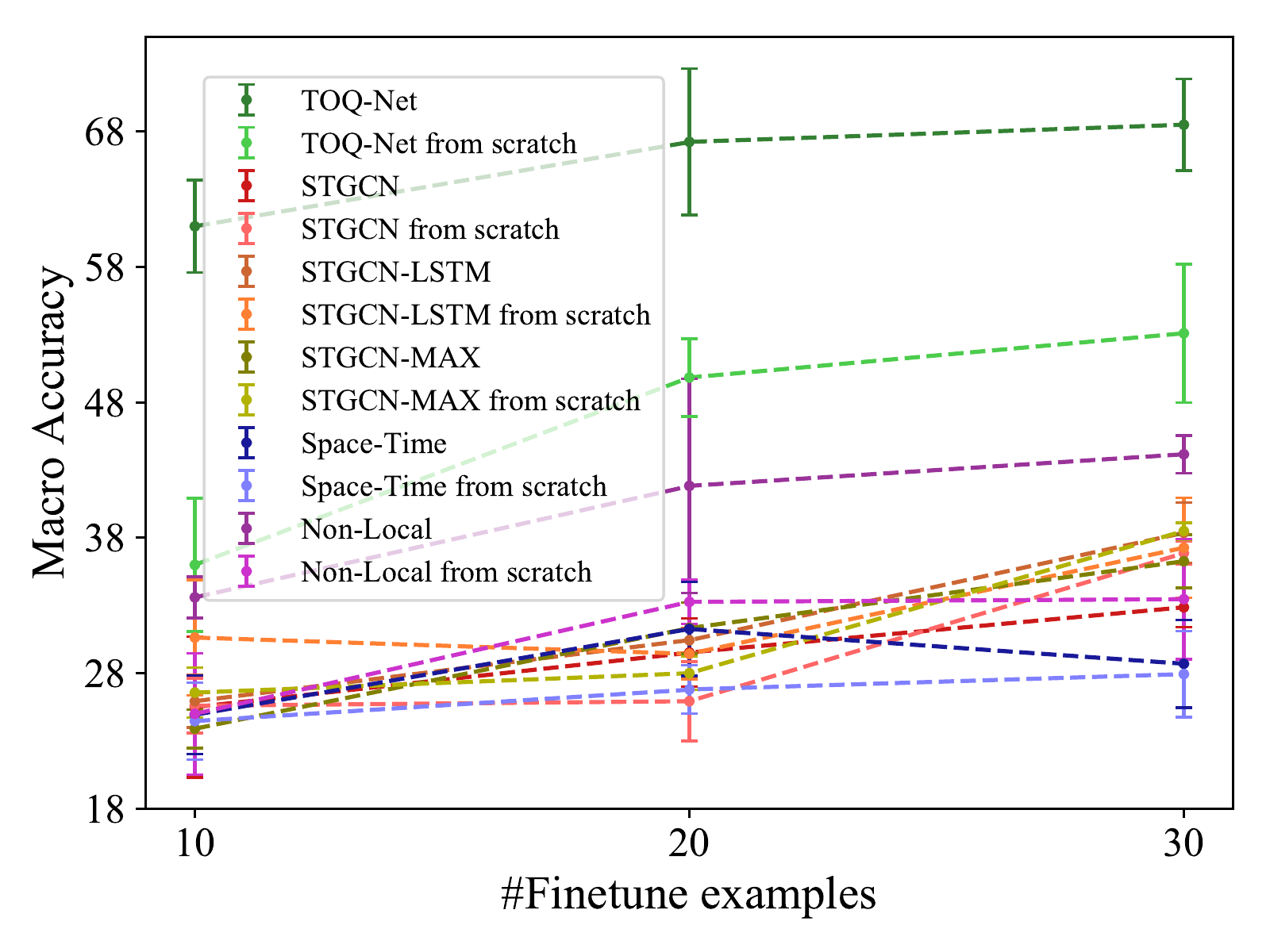}\captionof{figure}{Generalization to soccer environments with a different court size and agent speeds. The standard errors are computed based on three random seeds.}\label{fig:gfootball-2xinput}
\end{figure*}

Intuitively, after learning how to detect events in a soccer game, the learner should also generalize concepts to analogs with the same basic structure but enacted on proportionately larger or smaller spatial or temporal scales---as in the soccer-inspired game futsal. To evaluate generalization to different input scales, we first train all models on the original dataset with 9-way classification supervision. After that, we finetune each model on a new dataset where all input values are doubled (\ie, the court is now double the size, and the players now move at four times the speed), but with only a small number of examples. Thus, time flows twice fast, and the player moves four times the speed compared with the original dataset. Results are summarized in \fig{fig:gfootball-2xinput}. While all baselines can benefit from the pre-training on the original dataset, our \model outperforms all baselines by a significant margin when the finetuning dataset is small.

\subsection{Full Results for the Time-Warping Generalization}
In the paper we have evaluated the generalization performance of different models to time-warped sequences but with the same number of players. \tbl{tab:gfootball-generalization-full} shows the results on time-warped sequences with different number of agents.

\subsection{\#Layers of \model}

\begin{figure*}[t]
    \centering
    \vspace{-2pt}
    \includegraphics[width=0.5\textwidth]{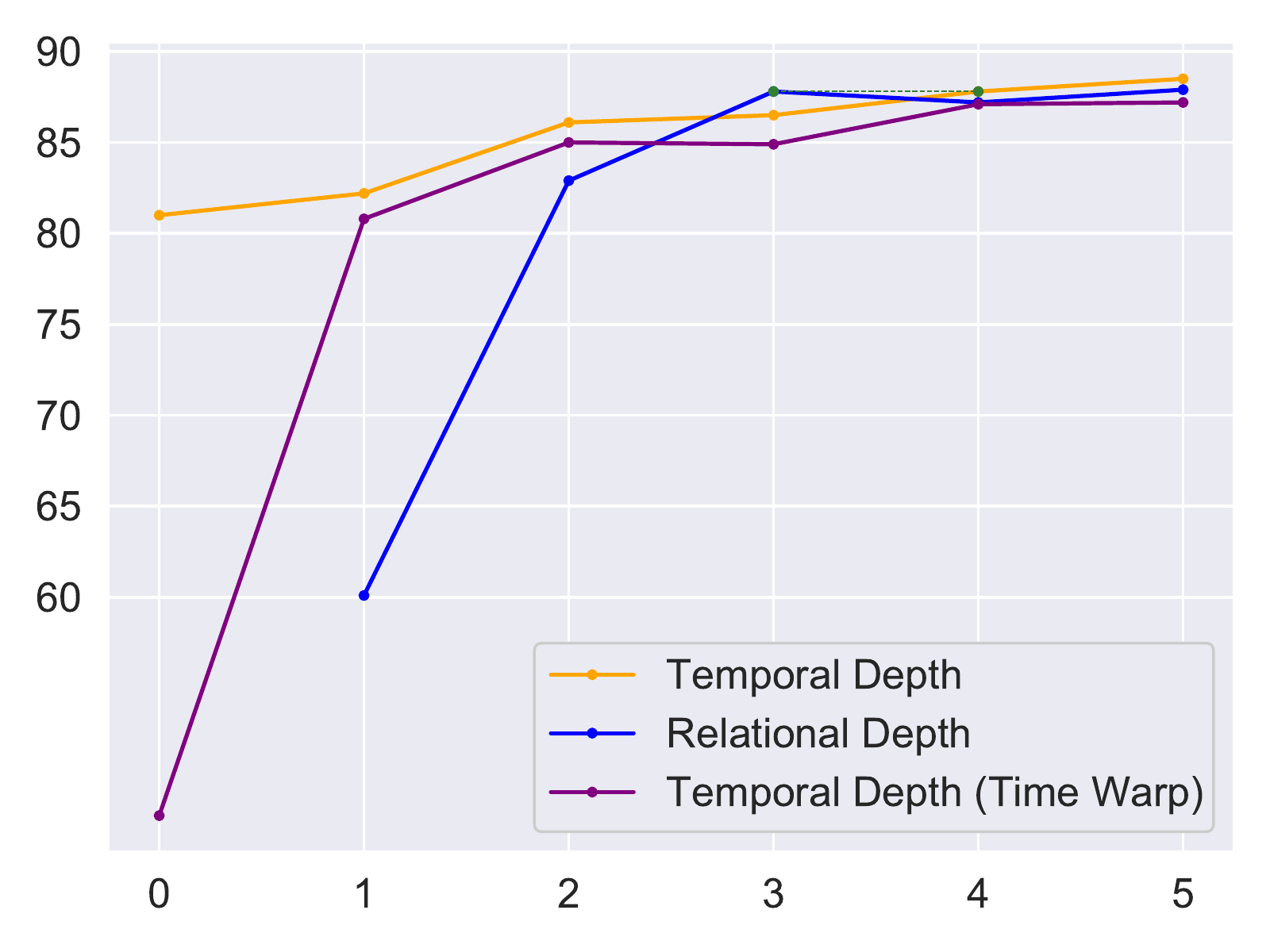}
    \vspace{-5pt}
    \captionof{figure}{Comparing \# of reasoning layers. Accuracy is tested on 6v6 9-way classification. When \# of relational reasoning layers vary (blue), temporal reasoning layers are fixed at 4. When \# of temporal reasoning layers vary (orange), relational reasoning layers are fixed at 3. When we have 0 temporal reasoning layers, we predict the sequence label based on the feature of the frame of interest. The purple line shows the performance on temporally warped trajectories when we have the \# of relational reasoning layers fixed and vary the \# of temporal reasoning layers. Relational layers are required for the prediction task (\# of relational reasoning layer must be greater than 0), because there is no nullary feature input to the network: we need at least one relational reasoning layer to gather information across the entities. Overall, adding reasoning layers improves results, and 3 relational layers + 4 temporal layers is a good balance between computation and performance.}
    \label{fig:layer_g}
\end{figure*}

We study the \model performance over varying number of relation reasoning layers and temporal reasoning layers on the 9-way classification task of the soccer event dataset (See figure \fig{fig:layer_g}), and decide to use the combination of 3 relational layers and 4 temporal layers. Both relational layers and temporal layers play important roles in the performance. As a generalization test, we also visualize the accuracy on temporally warped trajectories. We can see that temporal reasoning layers are important for the robustness against temporal warping.

\subsection{Interpretability}
\begin{figure*}[t]
    \centering
    \includegraphics[width=\textwidth]{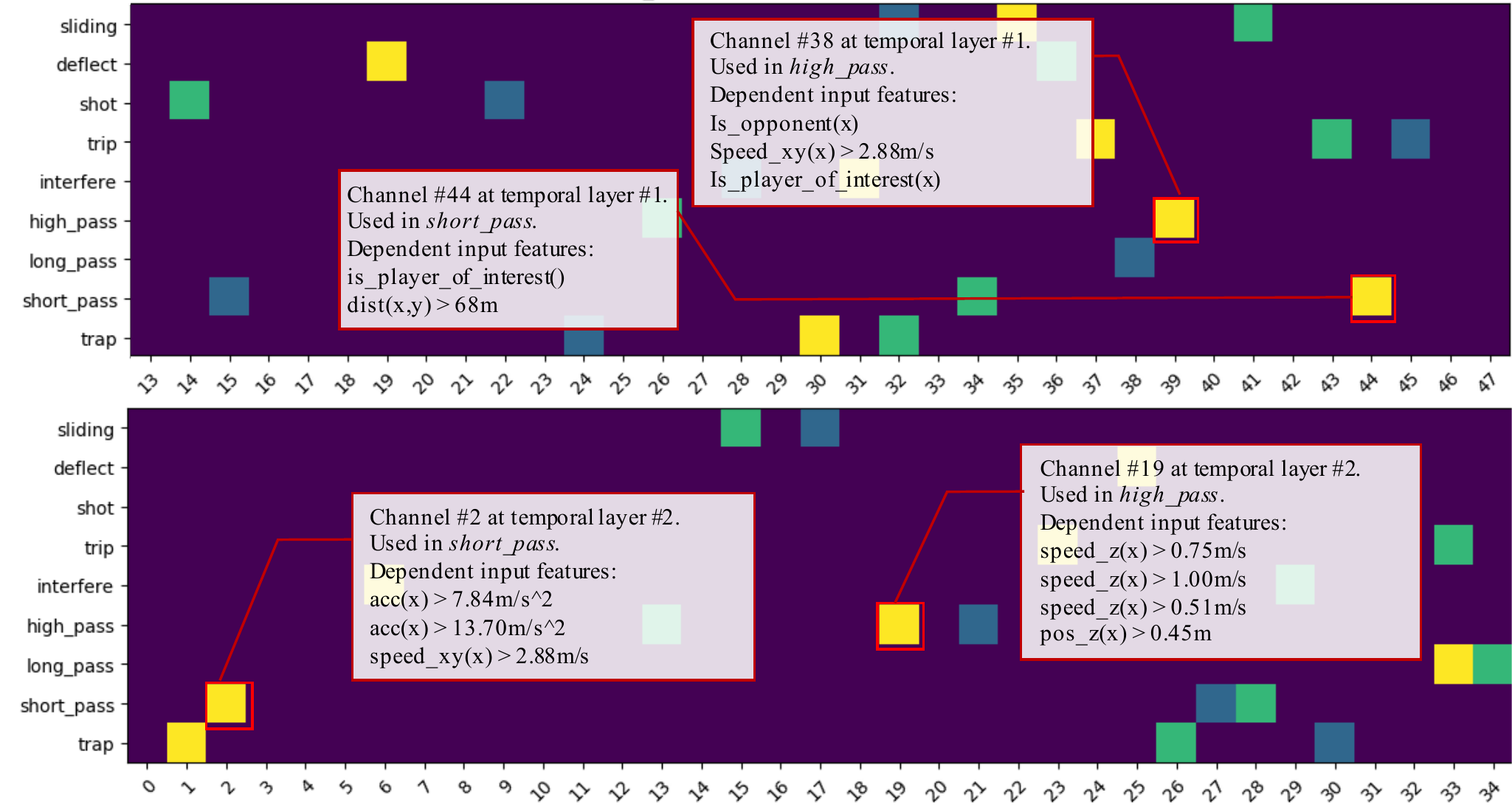}
    \vspace{-15pt}
    \captionof{figure}{Relevant features in temporal layers. Feature dependencies are computed by gradient. These dependencies and thresholds are learned end-to-end from data. Insets detail features for events in the first and the second stage of the {\it high pass} and {\it short pass}.}
    \label{fig:interpretability}
\end{figure*}

Though interpretability is not the main focus of our paper, in \fig{fig:interpretability}, we show how \model composes low-level concepts into high-level concepts: from inputs, to intermediate features at different temporal layers, and the final output labels. See the figure caption for detailed analysis.

\begin{table*}[!tp]
\vspace{-1.5em}
    \centering\small
    \begin{tabular}{lcccccc}
    \toprule
        Model    & Time Warp  & 3v3 & 4v4 & 6v6 & 8v8 & 11v11  \\ \midrule
STGCN & N & \cellcolor{yellow!25}{\mycellc{40.7$_{\pm 1.0}$\\{\scriptsize (-40.4\%)}}}  & \cellcolor{yellow!25}{\mycellc{63.2$_{\pm 4.9}$\\{\scriptsize (-7.4\%)}}}  & {\mycellc{68.2$_{\pm 2.8}$\\{\scriptsize (0.0\%)}}}  & \cellcolor{yellow!25}{\mycellc{55.4$_{\pm 3.3}$\\{\scriptsize (-18.8\%)}}}  & \cellcolor{yellow!25}{\mycellc{44.4$_{\pm 2.1}$\\{\scriptsize (-34.9\%)}}} \\
STGCN  & Y & \cellcolor{purple!5}{\mycellc{32.6$_{\pm 2.8}$\\{\scriptsize (-52.3\%)}}}  & \cellcolor{purple!5}{\mycellc{50.2$_{\pm 4.4}$\\{\scriptsize (-26.5\%)}}}  & \cellcolor{blue!5}{\mycellc{52.8$_{\pm 7.0}$\\{\scriptsize (-22.6\%)}}}  & \cellcolor{purple!5}{\mycellc{43.2$_{\pm 4.9}$\\{\scriptsize (-36.7\%)}}}  & \cellcolor{purple!5}{\mycellc{34.0$_{\pm 3.7}$\\{\scriptsize (-50.2\%)}}} \\
STGCN-MAX  & N & \cellcolor{yellow!25}{\mycellc{47.4$_{\pm 3.2}$\\{\scriptsize (-33.7\%)}}}  & \cellcolor{yellow!25}{\mycellc{68.8$_{\pm 2.0}$\\{\scriptsize (-3.8\%)}}}  & {\mycellc{71.5$_{\pm 1.9}$\\{\scriptsize (0.0\%)}}}  & \cellcolor{yellow!25}{\mycellc{59.1$_{\pm 0.7}$\\{\scriptsize (-17.3\%)}}}  & \cellcolor{yellow!25}{\mycellc{45.6$_{\pm 2.5}$\\{\scriptsize (-36.2\%)}}} \\
STGCN-MAX  & Y & \cellcolor{purple!5}{\mycellc{37.5$_{\pm 7.2}$\\{\scriptsize (-47.5\%)}}}  & \cellcolor{purple!5}{\mycellc{52.3$_{\pm 4.2}$\\{\scriptsize (-26.9\%)}}}  & \cellcolor{blue!5}{\mycellc{56.5$_{\pm 4.5}$\\{\scriptsize (-21.0\%)}}}  & \cellcolor{purple!5}{\mycellc{46.6$_{\pm 3.7}$\\{\scriptsize (-34.8\%)}}}  & \cellcolor{purple!5}{\mycellc{36.9$_{\pm 1.9}$\\{\scriptsize (-48.4\%)}}} \\
STGCN-LSTM  & N & \cellcolor{yellow!25}{\mycellc{39.7$_{\pm 1.1}$\\{\scriptsize (-43.1\%)}}}  & \cellcolor{yellow!25}{\mycellc{60.4$_{\pm 0.2}$\\{\scriptsize (-13.5\%)}}}  & {\mycellc{69.8$_{\pm 0.1}$\\{\scriptsize (0.0\%)}}} & \cellcolor{yellow!25}{\mycellc{55.8$_{\pm 2.0}$\\{\scriptsize (-20.0\%)}}}  & \cellcolor{yellow!25}{\mycellc{44.1$_{\pm 0.7}$\\{\scriptsize (-36.8\%)}}} \\
STGCN-LSTM  & Y & \cellcolor{purple!5}{\mycellc{21.8$_{\pm 0.8}$\\{\scriptsize (-68.8\%)}}}  & \cellcolor{purple!5}{\mycellc{27.8$_{\pm 1.3}$\\{\scriptsize (-60.2\%)}}}  & \cellcolor{blue!5}{\mycellc{30.6$_{\pm 0.6}$\\{\scriptsize (-56.1\%)}}}  & \cellcolor{purple!5}{\mycellc{25.8$_{\pm 1.0}$\\{\scriptsize (-63.1\%)}}}  & \cellcolor{purple!5}{\mycellc{22.6$_{\pm 0.8}$\\{\scriptsize (-67.6\%)}}} \\
Space-Time  & N & \cellcolor{yellow!25}{\mycellc{29.0$_{\pm 1.6}$\\{\scriptsize (-60.4\%)}}}  & \cellcolor{yellow!25}{\mycellc{53.5$_{\pm 3.2}$\\{\scriptsize (-27.0\%)}}}  & {\mycellc{73.3$_{\pm 0.3}$\\{\scriptsize (0.0\%)}}}  &  \cellcolor{yellow!25}{\mycellc{33.9$_{\pm 2.8}$\\{\scriptsize (-53.7\%)}}}  & \cellcolor{yellow!25}{\mycellc{15.2$_{\pm 1.8}$\\{\scriptsize (-79.3\%)}}} \\
Space-Time  & Y & \cellcolor{purple!5}{\mycellc{29.7$_{\pm 3.1}$\\{\scriptsize (-59.5\%)}}}  & \cellcolor{purple!5}{\mycellc{51.6$_{\pm 2.5}$\\{\scriptsize (-29.6\%)}}}  & \cellcolor{blue!5}{\mycellc{70.7$_{\pm 0.3}$\\{\scriptsize (-3.5\%)}}}  & \cellcolor{purple!5}{\mycellc{33.8$_{\pm 2.2}$\\{\scriptsize (-53.9\%)}}}  & \cellcolor{purple!5}{\mycellc{14.9$_{\pm 1.6}$\\{\scriptsize (-79.7\%)}}} \\
Non-Local  & N & \cellcolor{yellow!25}{\mycellc{45.9$_{\pm 5.1}$\\{\scriptsize (-41.2\%)}}}  & \cellcolor{yellow!25}{\mycellc{70.7$_{\pm 5.3}$\\{\scriptsize (-9.5\%)}}}  & {\mycellc{78.1$_{\pm 5.8}$\\{\scriptsize (0.0\%)}}} & \cellcolor{yellow!25}{\mycellc{58.5$_{\pm 10.8}$\\{\scriptsize (-25.1\%)}}}  & \cellcolor{yellow!25}{\mycellc{41.8$_{\pm 13.6}$\\{\scriptsize (-46.5\%)}}} \\
Non-Local  & Y & \cellcolor{purple!5}{\mycellc{46.7$_{\pm 4.1}$\\{\scriptsize (-40.2\%)}}}  & \cellcolor{purple!5}{\mycellc{69.9$_{\pm 4.7}$\\{\scriptsize (-10.5\%)}}}  & \cellcolor{blue!5}{\mycellc{77.7$_{\pm 5.0}$\\{\scriptsize (-0.5\%)}}}  & \cellcolor{purple!5}{\mycellc{58.7$_{\pm 12.6}$\\{\scriptsize (-24.9\%)}}}  & \cellcolor{purple!5}{\mycellc{41.3$_{\pm 13.6}$\\{\scriptsize (-47.1\%)}}} \\
\model  & N & \cellcolor{yellow!25}{\mycellc{\bf 77.4$_{\pm 3.5}$\\{\scriptsize (-12.4\%)}}}  & \cellcolor{yellow!25}{\mycellc{\bf 88.3$_{\pm 0.7}$\\{\scriptsize (-0.0\%)}}}  & {\mycellc{\bf 88.4$_{\pm 0.6}$\\{\scriptsize (0.0\%)}}} & \cellcolor{yellow!25}{\mycellc{81.3$_{\pm 1.7}$\\{\scriptsize (-8.0\%)}}}  & \cellcolor{yellow!25}{\mycellc{\bf 77.1$_{\pm 1.7}$\\{\scriptsize (-12.8\%)}}} \\
\model  & Y & \cellcolor{purple!5}{\mycellc{76.0$_{\pm 2.6}$\\{\scriptsize (-14.0\%)}}}  & \cellcolor{purple!5}{\mycellc{87.7$_{\pm 1.4}$\\{\scriptsize (-0.8\%)}}}  & \cellcolor{blue!5}{\mycellc{\bf 86.9$_{\pm 0.4}$\\{\scriptsize (-1.7\%)}}}  & \cellcolor{purple!5}{\mycellc{80.3$_{\pm 1.1}$\\{\scriptsize (-9.1\%)}}}  & \cellcolor{purple!5}{\mycellc{74.9$_{\pm 2.3}$\\{\scriptsize (-15.2\%)}}} \\
    \bottomrule
    \end{tabular}
    \vspace{-5pt}
    \caption{Full results on generalization to scenarios with all combinations of \#agents and temporally warping on the soccer event dataset. The standard error of all values are smaller than 2.5\%, computed based on three random seeds.
    }
    \label{tab:gfootball-generalization-full}
    \vspace{-5pt}
\end{table*}

\clearpage
{
\bibliographystyle{named}
\bibliography{actiongrounding}
}